\documentclass[runningheads]{llncs}

\usepackage{eccv}

\usepackage{eccvabbrv}
\usepackage[utf8]{inputenc} 
\usepackage[T1]{fontenc}    
\usepackage{latexsym}
\usepackage{microtype}
\usepackage{booktabs}
\usepackage{graphicx}
\graphicspath{{figures/}}
\usepackage{enumitem}
\usepackage{mathabx,amsbsy,mathtools,etoolbox,mathbbol}
\usepackage{algorithmic}
\usepackage[linesnumbered,ruled,vlined]{algorithm2e}
\usepackage{acronym}
\usepackage{balance}
\usepackage{xspace}
\usepackage{setspace}
\usepackage[skip=3pt,font=small]{subcaption}
\usepackage[skip=3pt,font=small]{caption}
\usepackage{tabularx,colortbl,multirow,array,makecell}
\usepackage{overpic,wrapfig}
\usepackage{nicefrac}
\usepackage[misc]{ifsym}
\usepackage{siunitx} 
\usepackage{soul} 
\usepackage{pifont}
\usepackage{relsize}
\usepackage[export]{adjustbox} 

\makeatletter
\DeclareRobustCommand\onedot{\futurelet\@let@token\@onedot}
\def\@onedot{\ifx\@let@token.\else.\null\fi\xspace}
\def\eg{\emph{e.g}\onedot} 

\def\ie{\emph{i.e}\onedot}

\def\etc{\emph{etc}\onedot}

\def\etal{\emph{et al}\onedot}
\makeatother
\newcommand{\sota}{\text{state-of-the-art}\xspace}
\newcommand\supp{\textit{supplementary}\xspace}

\newcommand{\cmark}{\ding{51}}
\newcommand{\xmark}{\ding{55}}

\makeatletter
\renewcommand\paragraph{\@startsection{paragraph}{4}{\z@}%
                       {-6.5\p@ \@plus -4\p@ \@minus -4\p@}%
                       {-0.5em \@plus -0.22em \@minus -0.1em}%
                       {\normalfont\normalsize\bfseries}}
\makeatother

\graphicspath{{figures/}}

\acrodef{nlp}[NLP]{Natural Language Processing}
\acrodef{vos}[VOS]{Video Object Segmentation}
\acrodef{3dvl}[3D-VL]{3D vision-language}
\acrodef{3dqa}[3D-QA]{3D question answering}
\acrodef{2dvl}[2D-VL]{2D vision-language}
\acrodef{ovseg}[OV-Seg]{open-vocabulary semantic segmentation}
\newcommand{\model}{\textsc{GPS}\xspace}
\acrodef{model}[\textsc{GPS}]{Grounded Pre-training for Scenes}
\newcommand{\dataset}{\textsc{SceneVerse}\xspace}

\acrodef{gru}[GRU]{Gated Recurrent Unit}
\acrodef{mhsa}[MHSA]{Multi-Head Self-Attention}
\acrodef{mha}[MHA]{Multi-Head Attention}
\acrodef{llms}[LLMs]{Large Language Models}
\acrodef{llm}[LLM]{Large Language Model}
\acrodef{mlm}[MLM]{Mask Language Modeling}

\newcommand{\scannet}{{ScanNet}\xspace}
\newcommand{\arkitscene}{{ARKitScenes}\xspace}
\newcommand{\multiscan}{{MultiScan}\xspace}
\newcommand{\rscan}{{3RScan}\xspace}
\newcommand{\hmthreeD}{{HM3D}\xspace}
\newcommand{\structured}{{Structured3D}\xspace}
\newcommand{\procthor}{{ProcTHOR}\xspace}

\newcommand{\colorPOP}{black}

\newcommand{\sceneNumTotal}{\textcolor{\colorPOP}{$68,406$}\xspace}
\newcommand{\sceneNumRough}{\textcolor{\colorPOP}{$68$K}\xspace}

\newcommand{\NumTotalRough}{\textcolor{\colorPOP}{$2.5$M}\xspace}

\newcommand{\annoNumOur}{\textcolor{\colorPOP}{$96,863$}\xspace}
\newcommand{\annoNumOurRough}{\textcolor{\colorPOP}{$97$K}\xspace}

\newcommand{\NumOfInstRough}{\textcolor{\colorPOP}{$1.5$M}\xspace}
\newcommand{\NumOfRel}{\textcolor{\colorPOP}{$21$}\xspace}

\newcommand{\NumOfGenTemp}{\textcolor{\colorPOP}{$1$M}\xspace}
\newcommand{\NumOfGenLLM}{\textcolor{\colorPOP}{$1$M}\xspace}

\acrodef{ari}[ARI]{Adjusted Rand Index}
\acrodef{msc}[MSC]{Mean Segmentation Covering}
\acrodef{miou}[mIoU]{mean Intersection over Union}
\acrodef{mse}[MSE]{Mean Squared Error}

\acrodef{ucla}[UCLA]{University of California, Los Angeles}
\acrodef{pku}[PKU]{Peking University}
\acrodef{thu}[THU]{Tsinghua University}
\acrodef{bigai}[BIGAI]{Beijing Institute of General Artificial Intelligence}

\definecolor{scope}{RGB}{103,78,167}
\definecolor{scenecaption}{RGB}{230,95,41}
\definecolor{objcaption}{RGB}{47,110,186}
\definecolor{objrefer}{RGB}{105,52,156}
\definecolor{semantic}{RGB}{230,145,56}
\definecolor{type}{RGB}{153,0,0}
\definecolor{revision}{RGB}{0,0,255}
\definecolor{revision}{RGB}{0,0,0}
\definecolor{gblue}{HTML}{4285F4}
\definecolor{gred}{HTML}{DB4437}
\definecolor{ggreen}{HTML}{0F9D58}
\definecolor{vblue}{HTML}{2993ba}
\newcommand{\blue}[1]{\textcolor{vblue}{\textbf{#1}}}

\makeatletter
\renewcommand*{\@fnsymbol}[1]{\ensuremath{\ifcase#1\or *\or \dagger\or \ddagger\or
   \mathsection\or \mathparagraph\or \|\or **\or \dagger\dagger
   \or \ddagger\ddagger \else\@ctrerr\fi}}
\makeatother

\usepackage{amsmath,amsfonts,bm}

\def\eqref#1{equation~\ref{#1}}

\def\1{\bm{1}}

\def\vb{{\bm{b}}}

\def\vf{{\bm{f}}}
\def\vg{{\bm{g}}}
\def\vh{{\bm{h}}}

\def\vl{{\bm{l}}}
\def\vm{{\bm{m}}}

\def\vo{{\bm{o}}}
\def\vp{{\bm{p}}}

\def\mT{{\bm{T}}}

\DeclareMathAlphabet{\mathsfit}{\encodingdefault}{\sfdefault}{m}{sl}
\SetMathAlphabet{\mathsfit}{bold}{\encodingdefault}{\sfdefault}{bx}{n}

\usepackage[accsupp]{axessibility}
\usepackage{hyperref}
\usepackage{orcidlink}

\begin{document}

\title{\textsc{\dataset}: Scaling 3D Vision-Language Learning for Grounded Scene Understanding}

\titlerunning{Scaling 3D Vision-Language Learning for Grounded Scene Understanding}

\author{Baoxiong Jia$^\star$\orcidlink{0000-0002-4968-3290} \and
Yixin Chen$^\star$\orcidlink{0000-0002-8176-0241} \and
Huangyue Yu\orcidlink{0009-0007-5729-0255} \and
Yan Wang\orcidlink{0009-0000-7001-3569} \and
Xuesong Niu\orcidlink{0000-0001-7737-4287} \and
Tengyu Liu\orcidlink{0000-0003-4006-1740}\and
Qing Li\orcidlink{0000-0003-1185-5365} \and
Siyuan Huang\orcidlink{0000-0003-1524-7148}
}

\authorrunning{B. Jia and Y. Chen et al.}

\institute{
$^\star$ indicates equal contribution\\
State Key Laboratory of General Artificial Intelligence, BIGAI\\
\url{https://scene-verse.github.io}
}
\maketitle
\begin{abstract}
3D vision-language (\acs{3dvl}) grounding, which aims to align language with 3D physical environments, stands as a cornerstone in developing embodied agents. In comparison to recent advancements in the 2D domain, grounding language in 3D scenes faces two significant challenges: (i) the scarcity of paired \acs{3dvl} data to support grounded learning of 3D scenes, especially considering complexities within diverse object configurations, rich attributes, and intricate relationships; and (ii) the absence of a unified learning framework to distill knowledge from grounded 3D data. In this work, we aim to address these major challenges in 3D-VL by examining the potential of systematically upscaling 3D-VL learning in indoor scenes. We introduce the first \textbf{million-scale} 3D-VL dataset, \dataset, encompassing \sceneNumRough indoor scenes and comprising \NumTotalRough vision-language pairs collected from both human annotations and our scalable scene-graph-based generation approach. We demonstrate that this scaling allows for a unified pre-training framework, Grounded Pre-training for Scenes (\model), for 3D-VL learning. Through extensive experiments, we showcase the effectiveness of \model by achieving \sota performance on existing 3D visual grounding and question-answering benchmarks. We also show that the data scaling effect is not limited to \model, but is generally beneficial for models on tasks like 3D semantic segmentation. The vast potential of \dataset and \model is unveiled through zero-shot transfer experiments in challenging \acs{3dvl} tasks.
\end{abstract}

\section{Introduction}\label{sec:intro}
The foundation of human cognitive development lies in the grounding of language within the physical world \cite{smith2005development, lake2017building, zhu2020dark}. Recent progress in \ac{llms}~\cite{brown2020language, touvron2023llama, bommasani2021opportunities} has markedly promoted the alignment between vision and language~\cite{radford2021learning,alayrac2022flamingo, liu2023visual} utilizing billion-scale vision-language datasets~\cite{schuhmann2022laion,zhu2023multimodal}. However, with these advancements predominantly focusing on the 2D domain, the grounded understanding of 3D physical environments remains in an incipient stage~\cite{chen2020scanrefer, achlioptas2020referit3d, azuma2022scanqa}. Recognizing the pivotal role of grounded 3D experiences in shaping human cognition~\cite{barsalou1999perceptual, barsalou2008grounded}, there is a compelling need to focus on exploring vision-language learning in the context of 3D scenes.

\begin{figure}[t]
    \centering
    \includegraphics[width=\linewidth]{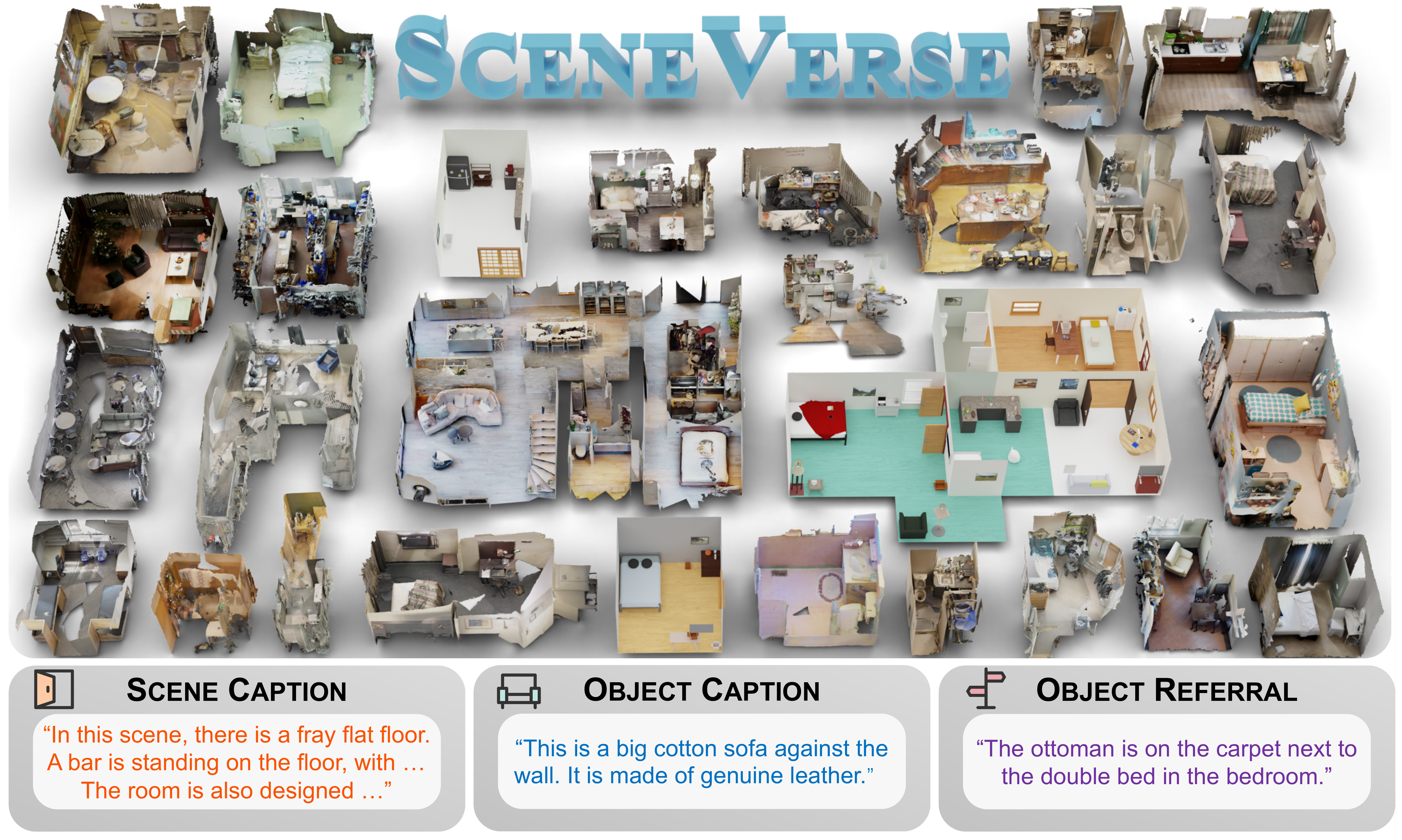}
    \caption{
        \textbf{Overview of \dataset.} A million-scale 3D vision-language dataset that comprises over \sceneNumRough various 3D indoor scenes and \NumTotalRough aligned scene-language pairs in the form of \textcolor{scenecaption}{scene caption}, \textcolor{objcaption}{object caption}, and \textcolor{objrefer}{object referral}.
    }
    \vspace{-0.25in}
    \label{fig:illustration}
\end{figure}%

Seeking insights from success in \ac{2dvl} learning, a major factor to the success was the notable scale-up of paired vision-language data~\cite{krishna2017visual, changpinyo2021conceptual, schuhmann2022laion}. However, applying this experience directly from 2D to 3D is fraught with challenges. Primarily, 3D data collection heavily relies on the scanning device, making it inherently much more complex and expensive than gathering 2D images. Despite steady efforts to increase the volume of 3D scene data~\cite{dai2017scannet,yeshwanth2023scannet++,mao2022multiscan,baruch2021arkitscenes}, most datasets remain limited to thousands of scenes, substantially lagging behind the scale of existing 2D datasets. This gap is further widened by the inherent complexities of 3D scenes, which feature a multitude of object instances with diverse attributes, varying arrangements, and intricate inter-object relationships. These unique aspects of 3D scenes not only make the accurate description of objects and their relations more challenging but also considerably increase the number of language descriptions required for thorough scene depiction. Consequently, this presents a significant challenge in gathering sufficient and high-quality paired scene-language data for grounded scene understanding.

To confront these challenges, we propose \dataset, the first \textbf{million-scale} dataset aimed at advancing \ac{3dvl} learning for grounded scene understanding. At the scene level, we unify 3D scene data from existing datasets~\cite{dai2017scannet,wald2019rio,baruch2021arkitscenes,ramakrishnan2021habitat,mao2022multiscan}, aligning scenes and annotations from various capturing sources, and supplement the collection with synthetic scenes~\cite{zheng2020structured3d,deitke2022️}. This compilation represents the most extensive 3D scene data gathered to date, amounting to \sceneNumRough scenes. For language, we first present \annoNumOurRough newly-annotated referring expressions, the most extensive thus far. We additionally propose an automated generation pipeline utilizing 3D scene graphs~\cite{armeni20193d, wald2020learning} and \ac{llms} to create comprehensive, high-quality scene-language pairs. This refined collection, totaling \NumTotalRough scene-language pairs, provides detailed and comprehensive descriptions of both object-level and scene-level descriptions within the 3D scene.

We thoroughly investigate the potential offered by \dataset with large-scale pre-training, introducing \ac{model}, a novel and unified pre-training framework designed for scene and object-level alignment without auxiliary losses. Through multi-level contrastive alignment, we achieve significant performance boosts on \ac{3dvl} tasks, such as grounding and question answering, setting new \sota results via a simple and effective pre-training process. We unveil the vast possibilities offered by \dataset and \model in \ac{3dvl} tasks in a zero-shot transfer setting. Additionally, we show that the scaling effect in \dataset is not limited to \model, but generally benefits models in tasks like 3D semantic segmentation. At last, we offer deeper insights into the data-scaling in \dataset through extensive ablative experiments, pointing out future directions.
Our main contributions are as follows:
\begin{enumerate}[nolistsep,noitemsep]
\item We introduce \dataset, the first million-scale \ac{3dvl} dataset for grounded scene understanding. \dataset encompasses \sceneNumRough 3D scenes coupled with \NumTotalRough scene-language pairs, sourced through a combination of human annotation and automated generation methods. This represents a significant improvement in terms of data diversity and scale compared to prior datasets.
\item We propose \model, a transformer-based model trained with multi-level scene-text alignment that achieves \sota results on existing \ac{3dvl} grounding and question-answering benchmarks by pre-training on \dataset.
\item We demonstrate that with the data scale-up and model design, our pre-trained \model exhibit emerging zero-shot generalization capabilities in grounded scene understanding. We also show that this scaling effect is not limited to \model, but is generally beneficial for models on tasks like semantic segmentation.
\end{enumerate}

\section{Related Work}\label{sec:related_work}
\paragraph{Datasets for Grounded 3D Understanding} Obtaining aligned 3D-language data is a inherently difficult task. In 3D object modeling, pioneering works like ShapeNet~\cite{chang2015shapenet} sourced 3D assets from online repositories, leading to a proliferation of high-quality 3D object datasets~\cite{mo2019partnet, collins2022abo,wu2023omniobject3d}. Notably, recent developments include internet-scale data collection with Objaverse~\cite{deitke2023objaverse,deitke2023objaversexl}, accompanied by the integration of object-level captions~\cite{xue2023ulip} for 3D-language alignment. Models trained on these datasets demonstrate an enhanced understanding of objects, evident in classification~\cite{liu2023openshape}, generation~\cite{liu2023zero}, and captioning tasks~\cite{luo2023scalable}.

In contrast, developing datasets for grounded 3D scene understanding is even more challenging due to the extensive requirements for scene acquisition and annotation. Existing works curate RGB-D and scanned indoor scene datasets~\cite{dai2017scannet,chang2017matterport3d,wald2019rio,baruch2021arkitscenes,ramakrishnan2021habitat,mao2022multiscan} and synthetic scenes~\cite{deitke2022️, zheng2020structured3d, khanna2024habitat, yang2024physcene} used for benchmarking tasks like 3D object detection and segmentation~\cite{ding2019votenet,jiang2020pointgroup,misra2021end,vu2022softgroup,schult2023mask3d}. These semantically labeled scenes are subsequently used in fine-grained scene grounding tasks like object referral~\cite{chen2020scanrefer,achlioptas2020referit3d,zhang2023multi3drefer}, captioning~\cite{chen2021scan2cap,yuan2022x,chen2021d3net,chen2023end}, vision-language-navigation~\cite{ma2019self,wang2019reinforced,pashevich2021episodic,hong2021vln} and reasoning~\cite{azuma2022scanqa,ma2022sqa3d,hong20233d}. Recent works exploit the representation of 3D scene graph (3DSG)~\cite{chen2019holistic++,armeni20193d, wald2020learning,rosinol2021kimera}, which concisely describes scenes with hierarchical structures. This representation is notably advantageous for planning~\cite{agia2022taskography,rana2023sayplan} and captioning~\cite{gu2023conceptgraphs}, owing to its compatibility with \ac{llms} for flexible description generation~\cite{hong20233d,huang2023embodied}. Nonetheless, as shown in~\cref{tab:stats_comparison}, most datasets are constrained in both scene and language scales, underscoring the need for scaling up fine-grained and aligned scene-language data to enhance grounded scene understanding.

\paragraph{Vision-Language Learning} Recent years have witnessed tremendous progress in \ac{2dvl}~\cite{radford2021learning,alayrac2022flamingo,li2022blip,liu2023visual,dai2023instructblip,chen2021yourefit}, empowered by transformer-based pre-training models~\cite{devlin2018bert,brown2020language,openai2023gpt4} and large-scale image-language datasets~\cite{changpinyo2021conceptual,schuhmann2022laion}. A central theme across \ac{2dvl} domains is the effectiveness of data scaling~\cite{kaplan2020scaling}, as demonstrated by improved alignment and expanded capabilities in open-vocabulary understanding~\cite{li2022grounded,li2022languagedriven,ghiasi2022scaling,kirillov2023segment} through a contrastive pre-training pipeline~\cite{radford2021learning}.

However, in grounded scene understanding, the primary challenge for models has been the limited availability of paired 3D scene-language data, which restricts the application of insights drawn from \ac{2dvl}. Current models for 3D scene grounding~\cite{he2021transrefer3d,zhao20213dvg,yang2021sat,bakr2022look,luo20223d,huang2022multi,jain2022bottom,chen2022language,wu2023eda} heavily rely on task-specific knowledge in both model and loss designs or advanced optimization strategies~\cite{zhu20233d}. To bridge this gap, there has been a growing emphasis on employing pre-trained \ac{2dvl} models for \ac{3dvl}~\cite{zhang2022pointclip,ha2022semantic,xue2023ulip,hegde2023clip,zhang2023learning,peng2023openscene,takmaz2023openmask3d}. Yet, these models mostly draw information available from \ac{2dvl} models (\eg, object attribute, affordance, \etc), falling short on capturing crucial 3D information like object spatial relationships which are necessary for more fine-grained tasks such as grounded human-scene~\cite{jiang2024scaling,wang2024move,huang2023diffusion,jiang2023full,wang2022humanise,chen2019holistic++} and robot-scene interactions modeling~\cite{mu2021maniskill,li2023behavior,mittal2023orbit,gong2023arnold}. This urges the need for a multi-level alignment between language and 3D scenes, particularly regarding 3D-specific information. Considering the nascent stage of existing 3D pre-training methods~\cite{ding2023pla,yang2023regionplc,zhu20233d,zhu2024unifying}, we believe \dataset and \model have the potential to spearhead new avenues in \ac{3dvl} research.

\begin{table}[t!]
    \small
    \centering
    \caption{\textbf{Comparison of \dataset with existing 3DVL Datasets.} \dataset expands the data scale of prior work by order of magnitude. ``VG'' stands for Visual Grounding, ``QA'' for Question Answering, ``PT'' for Pre-training and ``MT'' for Multi-tasking. ``Anno.'' denotes language from human annotations and ``Syn.'' for template-based or \acs{llm} generated descriptions.}
    \vspace{-10pt}
    \resizebox{\linewidth}{!}{
        \begin{tabular}{l|cc|c|ccc|c|cc|cc|c}
        \toprule
        \multirow{2}{*}{Dataset} & \multicolumn{2}{c|}{3D} & \multirow{2}{*}{Task} & Obj. & Scene & Obj. & Quality & \multicolumn{2}{c|}{New} & \multicolumn{2}{c|}{Existing} & \multirow{2}{*}{Total} \\
         & Scene & Obj.& & Caption & Caption & Referral & Check & Anno. & Syn. & Anno. & Syn. & \\
        \midrule
        ScanRefer~\cite{chen2020scanrefer}          & \multirow{2}[1]{*}{$|$} & \multirow{2}[1]{*}{$|$} & VG & \xmark & \xmark & \cmark & \cmark & 52K  & -   & - & - & 52K  \\
        ReferIt3D~\cite{achlioptas2020referit3d}    &      &  & VG & \xmark & \xmark & \cmark & \cmark   & 42K  & 200K & - & - & 242K \\
        ScanQA~\cite{azuma2022scanqa}               & 1.5K & 33K & QA  & - & - & - & \cmark  & 27K & - & - & - & 27K \\
        SQA3D~\cite{ma2022sqa3d}                    & \multirow{2}[1]{*}{$|$} & \multirow{2}[1]{*}{$|$} & QA & - & - & - & \cmark  & 33K & - & - & - & 33K \\
        Multi3DRefer~\cite{zhang2023multi3drefer}   &      &     & VG & \xmark & \xmark & \cmark  & \cmark  & - & 10K  & 52K & - & 62K  \\
        Cap3D\cite{luo2023scalable}               & -    & 666K & VG & \xmark & \cmark & \xmark & \xmark  & 58K  & 666K & - & - & 724K \\
        ScanScribe~\cite{zhu20233d}                & 3K   & 56K & PT & \xmark & \xmark & \cmark & \xmark  & -  & 90K & 94K & 94K & 278K \\
        3D-LLM~\cite{hong20233d} & 1.5K & 186K & MT & \cmark & \cmark & \cmark & \xmark  & -  & 659K & - & - & 659K \\
        EmbodiedScan~\cite{wang2023embodiedscan} & 5K & 890K & VG & \xmark & \xmark & \cmark & \xmark & - & 970K & - & - & 970K\\
        LEO~\cite{huang2023embodied} & 3K & 56K & MT & \cmark & \cmark & \cmark & \cmark  & -  & 188K & 235K & 90K & 513K \\
        \midrule
        \rowcolor[gray]{0.95} \blue{\dataset}              & \blue{68K}  & \blue{1.5M} & VG & \cmark & \cmark & \cmark & \cmark  & \blue{96K} & \blue{2.1M} & \blue{94K} & \blue{200K} & \blue{2.5M}   \\
        \bottomrule
        \end{tabular}
    }
    \label{tab:stats_comparison}
    \vspace{-0.2in}
\end{table}

\section{\texorpdfstring{\dataset}{}}\label{sec:data}
\dataset is designed for grounded scene understanding with 3D scenes curated from diverse existing datasets of both real and synthetic environments. Regarding language, we employ both human annotation and a novel automated generation pipeline to collect comprehensive and high-quality language for both object-level and scene-level descriptions. We provide details regarding data collection in the following sections.

\subsection{Scene Curation}
To address the scarcity of available 3D scene data, we construct \dataset by unifying 3D scene data from various existing datasets. We use real-world scene datasets, including \scannet~\cite{dai2017scannet}, \arkitscene~\cite{baruch2021arkitscenes}, \hmthreeD~\cite{ramakrishnan2021habitat}, \rscan~\cite{wald2019rio} and \multiscan~\cite{mao2022multiscan}, alongside synthetic environments from \structured~\cite{zheng2020structured3d} and \procthor~\cite{deitke2022️}. The inclusion of these synthetic datasets is mainly motivated by their potential as scalable data sources for \ac{3dvl} alignment.
To ensure cohesion across various sources, we conduct preprocessing steps such as room segmentation, point subsampling, axis alignment, normalization, and semantic label alignment. Each scan is represented by a point cloud $\mathrm{P} \in \mathbb{R}^{N\times8}$, wherein each point is defined by its 3D coordinates, RGB color, instance id, and semantic label. In total, we curate \sceneNumTotal 3D scenes in \dataset. 

\subsection{Referral Annotation by Humans}
In the curated scenes of \dataset, we present the most comprehensive set of human-annotated, context-rich object referrals to date, serving as a valuable benchmark for assessing grounded scene understanding capabilities. The human annotations contain \annoNumOur descriptions in \arkitscene~\cite{baruch2021arkitscenes}, \hmthreeD~\cite{ramakrishnan2021habitat} and \multiscan~\cite{mao2022multiscan}. During the annotation process, one human annotator was assigned to write at least 20 words to distinctly refer to a single 3D object within a 3D scene. Each referral text then undergoes independent verification by two additional reviewers, both mandated to accurately locate the referenced object based on the 3D scene and the annotated referral text. Any object referrals that do not pass the verification by either reviewer are flagged for re-annotation.

\begin{figure*}[t]
\centering
\includegraphics[width=\linewidth]{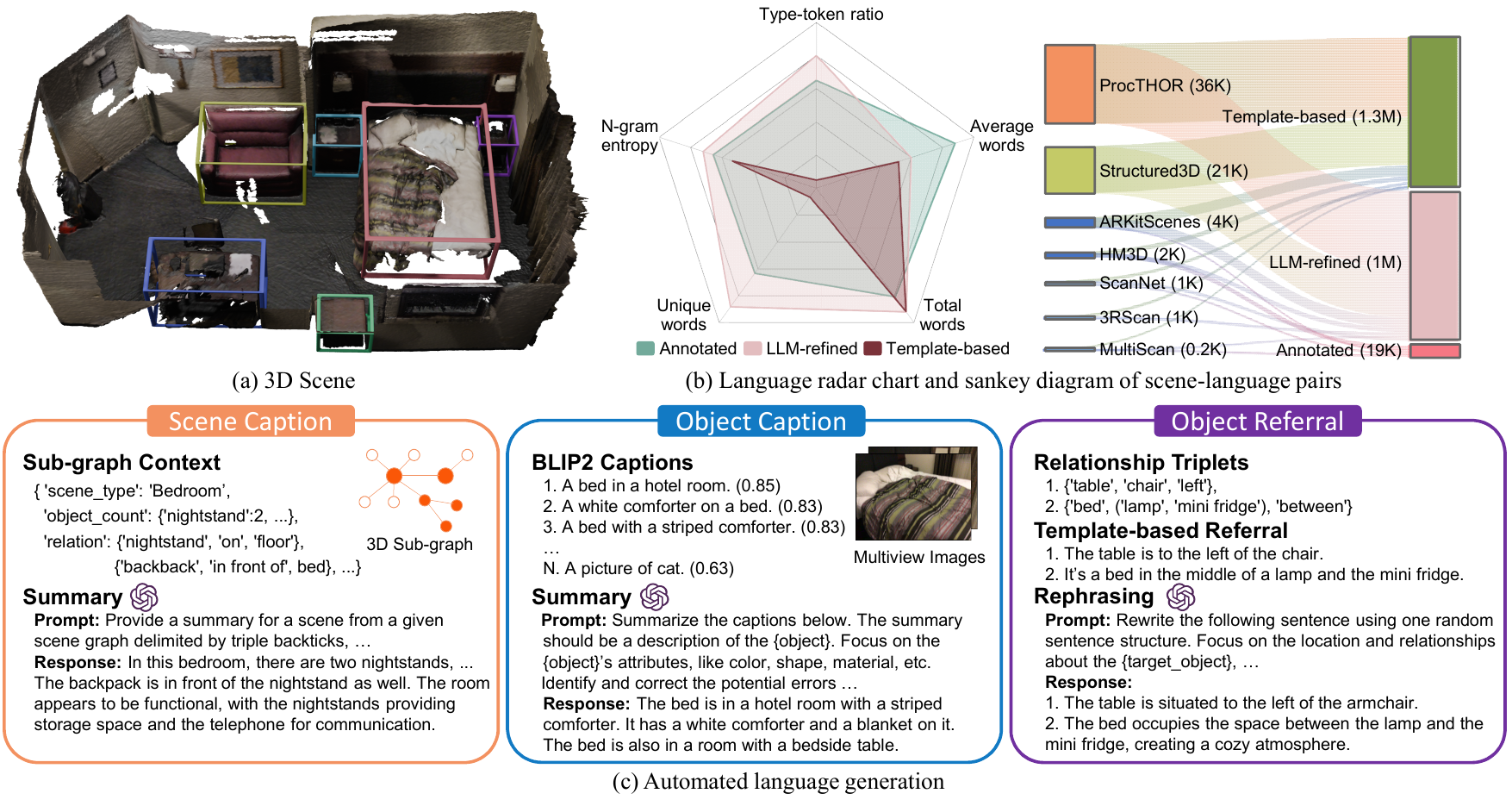}
\caption{\textbf{\dataset collection and statistics.} Given a 3D scene (a), our automated pipeline (c) generates three types of description including \textcolor{scenecaption}{scene caption}, \textcolor{objcaption}{object caption} and \textcolor{objrefer}{object referral}. (b) \dataset data comparison and composition.}
\label{fig:stats}
\vspace{-0.2in}
\end{figure*}
\subsection{3D Scene Graph Construction}\label{sec:data_scene_graph}
Our 3D scene graph is defined as a hierarchical graph $\mathcal{G}=(\mathcal{V},\mathcal{E})$. Each node $v\in\mathcal{V}$ represents one distinct 3D object instance, parameterized by its centroid $\vp_i \in \mathbb{R}^3$ and bounding box size of $\vb_i=(b_x,b_y,b_z) \in \mathbb{R}^3$. The edges $\mathcal{E}$ represent spatial relationships between nodes. To construct the scene graph $\mathcal{G}$, we instantiate the nodes with the instance annotation from the point clouds and assign object classes with their corresponding semantic labels. Following prior work\cite{achlioptas2020referit3d,wald2020learning}, we consider the \textbf{Vertical proximity}, \textbf{Horizontal proximity} and \textbf{Multi-object Relationships} as spatial relations. For a more detailed description of the scene graph construction and relationship determination, please refer to \supp.

\subsection{Language Generation with LLMs}\label{sec:data_language}
The scene-language pairs in \dataset aim to capture varying aspects of the 3D scene, including detailed object attributes in object captioning, spatial relationships between objects in object referral, and global scene descriptions in scene captioning. Based on 3D scene graphs, we utilize both templates and \ac{llms} to automatically generate descriptions on these three granularities. 

\paragraph{Object Captioning}
Object captions aim to provide detailed descriptions of an object's visual and physical properties, facilitating object-level grounding with its distinctive features. Given the multi-view images, we utilize the point cloud of the object $v \in \mathcal{V}$ to identify its occurrence in the images through rendering. The images are then cropped with the rendered bounding boxes and processed through BLIP2~\cite{li2023blip2} to generate initial object captions. We select the top 10 sentences with the highest CLIP~\cite{radford2021learning} similarity score and minimal occlusion and utilize an \acs{llm} to obtain a refined and coherent summary of the captions. 
The detailed object captioning pipeline is illustrated in \supp.

\paragraph{Object Referral}
Object relationship captions refer to objects by articulating their spatial relationships in the scene. Spatial relationship triplets $(v_i, v_j, e_{ij})$ are first extracted from the constructed 3D scene graph. We design various templates to generate descriptions for each relationship type, assigning the entities in the form of $(\text{\textcolor{ForestGreen}{target-object}}, \text{\textcolor{blue}{spatial-relation}}, \text{\textcolor{YellowOrange}{anchor-object(s)}})$. This results in examples like ``the \textcolor{ForestGreen}{chair} is \textcolor{blue}{next to} the \textcolor{YellowOrange}{armchair}'', ``facing the \textcolor{YellowOrange}{sofa}, there is a \textcolor{ForestGreen}{suitcase} \textcolor{blue}{far to the right of} the \text{\textcolor{YellowOrange}{shoes}}'', and ``the \textcolor{ForestGreen}{fridge} is \textcolor{blue}{between} \textcolor{YellowOrange}{cabinet} and \textcolor{YellowOrange}{sofa}''.
Our designed templates span passive and active tenses, as well as inversion clauses, contributing to the richness of the generated text. To enhance the descriptions' naturalness, we employ \acs{llm} for sentence rephrasing. Statistics for the descriptions before and after rephrasing are presented in \cref{fig:stats}~(b).

\paragraph{Scene Captioning}
The scene-level captions emphasize global information, portraying the key objects in the scene along with their attributes and functionalities. We use the constructed 3D scene graph and prompt \ac{llms} to generate these captions. We random sample a subset of edges and nodes from the scene graph each time as the scene context to enhance the diversity of scene captions. The object counts are also provided as \acs{llm} prompts, together with the room type and object attributes if such annotations are available in the dataset.

\subsection{Data Quality and Statistics}

\paragraph{Data Quality} The $96$K human-annotated set of \dataset is collected through AMT, where $82$ humans are employed for annotation and $18$ for verification. All final annotations passed the reference verification, with a re-annotation rate of $4.8\%$. For our automatic language generation pipeline, we conduct extensive prompt tuning and iterate with human feedback for \acs{llm}s on object captioning, summary, and rephrasing. To verify the efficacy of the pipeline, we conduct a quality check where $12$K generated object-level descriptions are randomly selected for human verification. Results demonstrate a $96.93\%$ pass rate, surpassing that in ReferIt3D~\cite{achlioptas2020referit3d} with $86.1\%$ pass rate on $2$K samples.

\paragraph{Statistics} In total, \dataset comprises a total of \sceneNumTotal room-level 3D scans, with the source composition shown in \cref{fig:stats}~(b). The dataset contains \NumOfInstRough object instances ranging in 2290 object categories. Our generated 3D scene graph comprises \NumOfRel types of relationships following prior work~\cite{achlioptas2020referit3d,wald2020learning}. For the language descriptions, we generate \NumOfGenTemp template-based texts and \NumOfGenLLM sentences rephrased by Llama~\cite{touvron2023llama} and GPT-3.5~\cite{openai2022chatgpt}. As can be seen from the radar chart and examples in \cref{fig:stats}, the diversity of the LLM-refined data, particularly in sentence length and variety, closely aligns with the characteristics of annotated descriptions, surpassing the template-based data. Together with existing sources (294K) and our newly annotated set (96K), \dataset contains \NumTotalRough scene-language pairs in total. All the rephrasing and summary prompts, along with the complete set of relationships, are detailed in \supp. 

\section{Grounded Pre-training for Scenes}\label{sec:model}

\begin{figure}[t]
    \includegraphics[width=\linewidth]{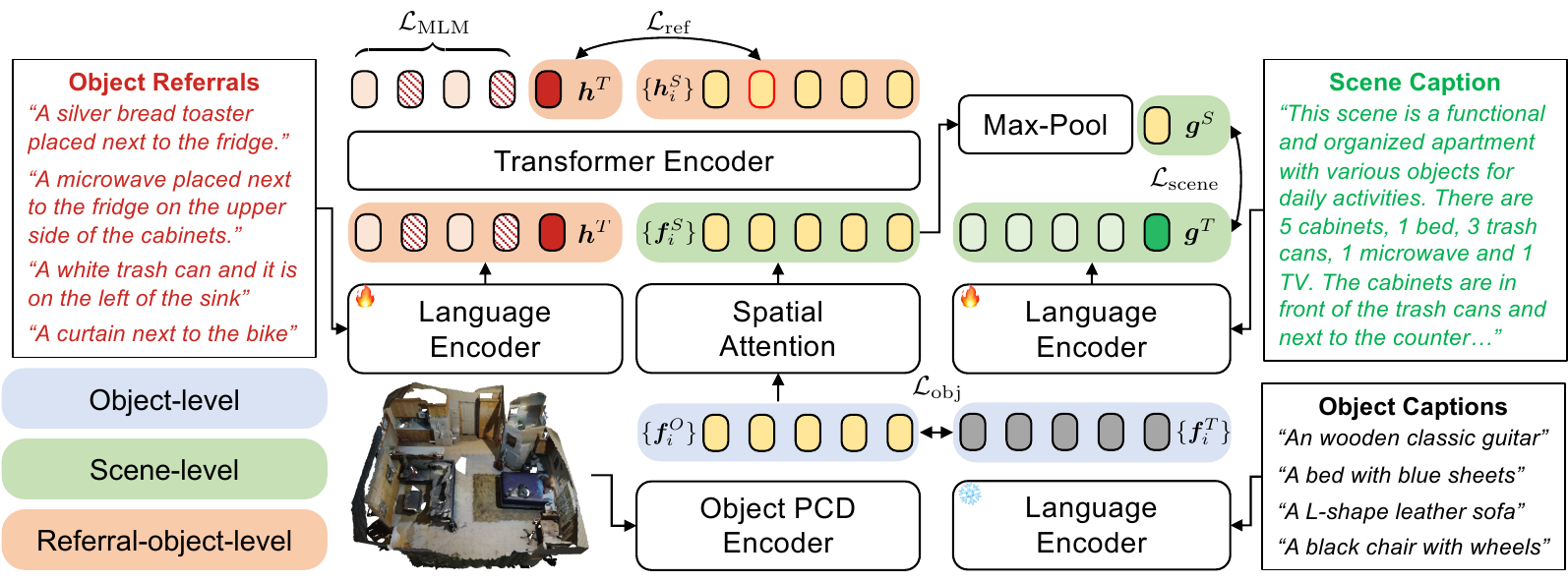} 
    \caption{\textbf{Overview of \model model.} We use contrastive alignment at three levels $\mathcal{L}_{\text{obj}}$, $\mathcal{L}_{\text{scene}}$, and $\mathcal{L}_{\text{ref}}$ and a masked language modeling objective $\mathcal{L}_{\text{MLM}}$ for model learning.}
    \label{fig:model}
    \vspace{-.2in}
\end{figure}
In this section, we introduce \ac{model}, an efficient transformer-based model trained with multi-level contrastive losses for aligning 3D scenes and texts. As shown in~\cref{fig:model}, we echo the language descriptions collected at different granularities to form contrastive objectives at both object-level, referral-object-level, and scene-level in \ac{model}. We describe the design of each level in the following sections.

\subsection{Object-level Grounding}\label{sec:model:object}
Given a 3D scene point cloud $\mathcal{S}$, we use an off-the-shelf 3D object segmentation model to decompose it into a bag of $N$ objects $\mathcal{S}=\left\{\vo_1, \vo_2, \cdots, \vo_n\right\}_{i=1}^N$. We extract object features $\{\vf^{O}_i\}$ with an object point cloud encoder and text features $\{\vf^{T}_i\}$ by feeding object-captions $\{\mT^{\text{obj}}_i\}$ into a frozen language model. Following~\cite{xue2023ulip}, we perform cross-modal alignment on the object features and text features via:
\begin{equation}
\begin{aligned}
    \mathcal{L}_{\text{obj}} = -\frac{1}{2}\sum_{(p,q)}  & \left( \log\frac{\exp{\left(D^{\text{obj}}(p,q)\right)}}{\sum_{r}\exp{\left(D^{\text{obj}}(p,r)\right)}} + \log\frac{\exp{\left(D^{\text{obj}}(p,q)\right)}}{\sum_{r}\exp{\left(D^{\text{obj}}(r,q)\right)}}\right),
\end{aligned}
\end{equation}
where $D^{\text{obj}}(p, q) = (\vf^{O}_p\vf^{T}_q / \tau)$ denotes the dot product between object and text features and $(p, q)$ denotes a pair of aligned object-text pair in the training batch and $r$ iterates over all object-text pairs in the training batch. Similar to CLIP~\cite{radford2021learning}, we use a learnable temperature parameter $\tau$ to facilitate model learning. 

\subsection{Scene-level Grounding}\label{sec:model:scene}
With aligned object features, we encode the scene by incorporating object spatial locations into the extracted object features. Specifically, we use a spatial transformer model to encode extracted object features $\{\vf^{O}_i\}$ with their spatial location features $\{\vl_i\}$ following~\cite{chen2022language,zhu20233d}:
\begin{equation*}
    \vf^{S} = \mathrm{SpatialAttn}(\{\vf_i^O\}, \{\vl_i\})
\end{equation*}
where $\{\vf_i^S\}$ denotes the feature of object $\vo_i$ after encoding with spatial location features.
To perform scene-level alignment, we operate on these scene-level object features $\{\vf_i^S\}$ and align it with the scene caption $\mT^{\text{scene}}$. Specifically, we feed the object features into a projection layer and use max-pooling over all object features to obtain the scene feature $\vg^{S}$. Similar to object-level grounding, we pass the scene caption through a tunable language model to obtain text feature $\vg^{T}$ and perform scene-level contrastive alignment through:
\begin{equation}
\begin{aligned}
    \mathcal{L}_{\text{scene}} = -\frac{1}{2}\sum_{(p,q)} & \left(\log\frac{\exp{\left(D^{\text{scene}}(p,q)\right)}}{\sum_{r}\exp{\left(D^{\text{scene}}(p,r)\right)}} + \log\frac{\exp{\left(D^{\text{scene}}(p,q)\right)}}{\sum_{r}\exp{\left(D^{\text{scene}}(r,q)\right)}}\right),
\end{aligned}
\end{equation}
where $D^{\text{scene}}(p,q)=(\vg_p^{S}\vg_q^{T} / \tau)$ denotes the dot product between scene feature $\vg_p^S$ and scene caption feature $\vg_q^{T}$ for each pair of aligned scene-text pairs in the training batch and $r$ iterates over all scene-text pairs in the training batch.

\subsection{Referral-object-level Grounding}\label{sec:model:refer}
To model the relationships revealed in referring expressions, we employ a self-attention-based reasoning transformer for grounding object referrals in scenes. This transformer takes in scene-object features $\{\vf^{S}_i\}$ and an object referral $\mT^{\text{ref}}$ and performs self-attention to learn relationships between text descriptions and object relationships. We use the same tunable language encoder as in scene-level grounding for extracting per-object referral features. We pass this text feature together with scene-object features into the self-attention transformer to obtain the aligned object features $\vh^{S}_{i}$ and the sentence-level referral feature $\vh^{T}$. We then perform the referral-object-level contrastive alignment following:
\begin{equation}
\begin{aligned}
    \mathcal{L}_{\text{ref}} = -\log\frac{\exp{\left(\bar{\vh}^S\vh^T/\tau\right)}}{\sum_{p}\exp{\left(\vh_p^S\vh^T/\tau\right)}},
\end{aligned}
\end{equation}
where $\bar{\vh}^S$ denotes the feature of the referred object, $p$ iterates over all objects within the same scene. Notably, in contrast to inter-scene contrast that was done in object- and scene-level alignment, we force the selection of positive pairs to be within the same scene to provide intra-scene contrast for fine-grained object grounding. This mimics the success of intra-image and inter-image contrasts commonly used for region-word alignment in \ac{2dvl} models~\cite{zhang2022glipv2}. 

To learn the multi-level alignment between 3D scenes and language, we first train the point cloud encoder with an object-level grounding objective to obtain a good feature initialization for grounding objects in scenes. During the scene grounding stage, we train our inter- and intra-scene objectives together with a masked language modeling loss $\mathcal{L}_{\text{MLM}}$ over the inputted object-referral texts to tune the parameters within the language encoder and self-attention transformer. Above all, the learning of \ac{model} could be summarized as optimizing:
\begin{equation*}
    \mathcal{L} = \mathcal{L}_{\text{obj}} + \mathcal{L}_{\text{scene}} + \mathcal{L}_{\text{ref}} + \mathcal{L}_{\text{MLM}}.
\end{equation*}

\section{Experiments}\label{sec:exp}
In this section, we present experimental results addressing the following questions:
\begin{enumerate}
\item How effective is the data scaling in \dataset for 3D visual grounding? Does the scale-up benefit common \ac{3dvl} tasks (\eg, 3D question answering, open-vocabulary 3D semantic segmentation) and pre-training-based models?

\item How well is the \model pre-training pipeline for \ac{3dvl} tasks? Does it exhibit similar properties of \ac{2dvl} models in \ac{3dvl} tasks?
\item What is offered by \dataset and \model and what is missing?
\end{enumerate}
In the following sections, we describe in detail the model performance regarding these key topics. Due to the page limit, we direct readers to the \supp for implementation details, qualitative results, and more experimental analyses.
\subsection{3D Visual Grounding}\label{sec:exp:grounding}
\paragraph{Settings} We evaluate our model on three commonly-used datasets for 3D visual grounding: ScanRefer~\cite{chen2020scanrefer}, Nr3D, and Sr3D~\cite{achlioptas2020referit3d}. For Nr3D and Sr3D, we follow Achlioptas \etal~\cite{achlioptas2020referit3d} and report the grounding accuracies of models using ground-truth object masks. For ScanRefer, we follow Zhu \etal~\cite{zhu20233d} and use Mask3D~\cite{schult2023mask3d} to generate object proposals. Results are reported as $\text{Acc}@0.5$ to evaluate the correctness of predictions whose object bounding boxes overlap the ground truth with IoU $> 0.5$. For comparisons, we compare with existing baselines by providing the results of pre-trained \ac{model} and dataset-specific fine-tuned \ac{model}. Please see more details in the \supp.
\paragraph{Results and Analyses} As shown in~\cref{tab:referit3d}, \ac{model} trained on \dataset achieves \sota results on all existing \ac{3dvl} grounding benchmarks. Initially, when \model is trained directly on the training sets of benchmark datasets, labeled as Ours (\textit{scratch}), it underperforms compared to existing models that employ more complex structures or loss designs. This result underscores the data-intensive nature of the contrastive alignment paradigm. However, when presented with extensive training data in \dataset, the results of our model without additional fine-tuning, \ie, Ours (\textit{pre-train}), significantly improves and already achieves \sota results on benchmarks like ScanRefer. Moreover, the dataset-specific fine-tuned model, \ie, Ours (\textit{fine-tuned}), \textit{consistently outperforms existing baselines with only a simple projection MLP} added on top of the pre-trained model, jointly optimized during fine-tuning without any other auxiliary architecture or loss objective. These results underscore the strong potential of both the \dataset and \model for \ac{3dvl} tasks. 

\begin{table*}[t!]
    \centering
    \caption{\textbf{3D visual grounding results on Nr3D, Sr3D, and ScanRefer.} We use ``\textit{pre-train}'' for our model trained on \dataset w/o additional fine-tuning, and ``\textit{fine-tune}'' for its data-specific fine-tuned version. Best results are highlighted in \textbf{bold}.}
    \vspace{-10pt}
    \resizebox{\linewidth}{!}{
        \begin{tabular}{lccccccccccccc}
        \toprule
        \multirow{2}[2]{*}{Method} & \multicolumn{5}{c}{Nr3D} & \multicolumn{5}{c}{Sr3D} & \multicolumn{3}{c}{ScanRefer Acc@0.5}\\
        \cmidrule(lr){2-6}\cmidrule(lr){7-11}\cmidrule(lr){12-14}
        & Overall & Easy & Hard & V-Dep. & V-Indep. & Overall & Easy & Hard & V-Dep. & V-Indep. & Overall & Unique & Multiple \\
        \midrule
        3DVG-Trans~\cite{zhao20213dvg} & 40.8 & 48.5 & 34.8 & 34.8 & 43.7 & 51.4 & 54.2 & 44.9 & 44.6 & 51.7 & 34.7 & 60.6 & 28.4 \\
        TGNN~\cite{huang2021text} & 37.3 & 44.2 & 30.6 & 35.8 & 38.0 & 45.0 & 48.5 & 36.9 & 45.8 & 45.0 & 29.7 & 56.8 & 23.2 \\
        TransRefer3D~\cite{he2021transrefer3d} & 48.0 & 56.7 & 39.6 & 42.5 & 50.7 & 57.4 & 60.5 & 50.2 & 49.9 & 57.7 & - & - & - \\
        InstanceRefer~\cite{yuan2021instancerefer} & 38.8 & 46.0 & 31.8 & 34.5 & 41.9 & 48.0 & 51.1 & 40.5 & 45.8 & 48.1 & 32.9 & 66.8 & 24.7 \\
        FFL-3DOG~\cite{feng2021free} & 41.7 & 48.2 & 35.0 & 37.1 & 44.7 & - & - & - & - & - & 34.0 & 67.9 & 25.7 \\
        LAR~\cite{bakr2022look} & 48.9 & 58.4 & 42.3 & 47.4 & 52.1 & 59.4 & 63.0 & 51.2 & 50.0 & 59.1 & - & - & - \\
        SAT~\cite{yang2021sat} & 56.5 & 64.9 & 48.4 & 54.4 & 57.6 & 57.9 & 61.2 & 50.0 & 49.2 & 58.3 & 30.1 & 50.8 & 25.2 \\
        3D-SPS~\cite{luo20223d} & 51.5 & 58.1 & 45.1 & 48.0 & 53.2 & 62.6 & 56.2 & 65.4 & 49.2 & 63.2 & 37.0 & 66.7 & 29.8 \\
        3DJCG~\cite{cai20223djcg} & - & - & - & - & - & - & - & - & - & - & 37.3 & 64.3 & 30.8 \\
        BUTD-DETR~\cite{jain2022bottom} & 54.6 & 60.7 & 48.4 & 46.0 & 58.0 & 67.0 & 68.6 & 63.2 & 53.0 & 67.6 & 39.8 & 66.3 & 35.1 \\
        MVT~\cite{huang2022multi} & 59.5 & 67.4 & 52.7 & 59.1 & 60.3 & 64.5 & 66.9 & 58.8 & 58.4 & 64.7 & 33.3 & 66.5 & 25.3 \\
        ViL3DRel~\cite{chen2022language} & 64.4 & 70.2 & 57.4 & \textbf{62.0} & 64.5 & 72.8 & 74.9 & 67.9 & 63.8 & 73.2 & 37.7 & 68.6 & 30.7 \\
        EDA~\cite{wu2023eda} & 52.1 & 58.2 & 46.1 & 50.2 & 53.1 & 68.1 & 70.3 & 62.9 & 54.1 & 68.7 & 42.3 & 68.6 & 37.6 \\
        3D-VisTA (\textit{scratch})~\cite{zhu20233d} & 57.5 & 65.9 & 49.4 & 53.7  & 59.4 & 69.6 & 72.1 & 63.6 & 57.9 & 70.1 & 41.5 & 70.9 & 34.8 \\ 
        3D-VisTA~\cite{zhu20233d} & 64.2 & 72.1 & 56.7 & 61.5 & 65.1 & 76.4 & 78.8 & 71.3 & 58.9 & 77.3 & 45.8 & 75.1 & 39.1 \\ 
        \midrule
        Ours (\textit{scratch}) & 58.7 & 67.0 & 50.9 & 55.8 & 59.8 & 68.4 & 70.5 & 63.4 & 53.1 & 69.0 & 40.4 & 71.3 & 34.7 \\ 
        Ours (\textit{pre-train}) & 55.2 & 62.8 & 48.0 & 45.5 & 58.8 & 74.1 & 76.4 & 68.5 & 54.1 & 75.0 & 47.1 & 77.4 & 41.6 \\ 
        Ours (\textit{fine-tuned}) & \textbf{64.9} & \textbf{72.5} & \textbf{57.8} & 56.9 & \textbf{67.9} & \textbf{77.5} & \textbf{80.1} & \textbf{71.6} & \textbf{62.8} & \textbf{78.2} & \textbf{48.1} & \textbf{77.9} & \textbf{42.7} \\
        \bottomrule
        \end{tabular}
    }
    \label{tab:referit3d}
    \vspace{-0.15in}
\end{table*}


\subsection{Zero-Shot Transfer}\label{sec:exp:zero-shot}
\paragraph{Settings} To better evaluate the effectiveness of both the \dataset data and the \model model, we further perform zero-shot transfer experiments to test the models' capability in 4 benchmarks, ScanRefer, Sr3D, Nr3D, and \dataset-val. We create \dataset-val using $8.5K$ annotated object referrals of $271$ scenes in MultiScan, and randomly split the scenes following a 4:1 train / test split for creating the held-out test set.
We mainly consider 2 specific transfer settings in our experiments: (i) \textit{zero-shot}: models trained by removing all the scenes from the target dataset, tested on held-out unseen scenes, and (ii) \textit{zero-shot text}: Models trained on data that include the training set of scenes from the target dataset, yet tested exclusively with unseen scene-text distribution. Specifically, for the \textit{zero-shot text} setting, we use the generated texts in \dataset as fine-tuning sources for the \textit{zero-shot} model. We mainly compare our model against a recent pre-training-based model 3D-VisTA. See more details on experimental setting and implementation in the \supp.

\begin{table}[t!]
    \begin{minipage}{\linewidth}
        \centering
        \begin{minipage}[t]{0.51\linewidth}
            \caption{\textbf{Zero-shot transfer on existing benchmarks.} ``SR'' stands for ScanRefer.}
            \vspace{-10pt}
            \resizebox{\linewidth}{!}{
                \begin{tabular}{lcccc}
                \toprule
                Method & Nr3D & Sr3D & SR@0.25 & SR@0.5\\
                \midrule
                3D-VisTA (\textit{scratch}) & 57.5 & 69.6  & 45.9 & 41.5\\
                3D-VisTA (\textit{zero-shot}) & 35.2 & 31.2 & 33.2 & 29.6 \\
                3D-VisTA (\textit{zero-shot text}) & 43.1 & 36.1 & 41.1 & 36.4 \\
                \midrule
                Ours (\textit{scratch}) & 58.7 & 68.4 & 44.5 & 40.4 \\
                Ours (\textit{zero-shot}) & 32.4 & 33.3 & 35.2 & 31.1  \\
                Ours (\textit{zero-shot text}) & 41.9 & 38.1 & 40.7 & 35.8 \\
                \bottomrule
                \end{tabular}
            }
            \label{tab:scannet_transfer}
        \end{minipage}
        \hfill
        \begin{minipage}[t]{0.465\linewidth}
            \centering
            \captionof{table}{\textbf{Zero-shot transfer on \dataset-val.} Evaluation uses GT object proposals following Nr3D/Sr3D.}
            \vspace{-10pt}
            \resizebox{\linewidth}{!}{
                \begin{tabular}{lccccc}
                \toprule
                Method & Overall & Easy & Hard & V-Dep. & V-Indep.\\
                \midrule
                3D-VisTA (\textit{scratch})  & 40.7 & 53.1 & 21.6 & 37.3 & 44.3 \\
                3D-VisTA (\textit{zero-shot})     & 52.9 & 59.6 & 35.4 & 53.7 & 52.2 \\
                3D-VisTA (\textit{zero-shot text})   & 58.1 & 70.0 & 39.6 & 52.5 & 64.1 \\
                \midrule
                Ours (\textit{scratch}) & 38.5 & 50.2 & 20.8 & 33.7 & 43.9 \\
                Ours (\textit{zero-shot}) & 59.2 & 69.4 & 44.0 & 53.1 & 66.3 \\
                Ours (\textit{zero-shot text}) & 60.6 & 70.9 & 45.1 & 54.8 & 67.3 \\
                \bottomrule
                \end{tabular}
            }
            \label{tab:multiscan_transfer}
        \end{minipage}
    \end{minipage}
    \vspace{-0.1in}
\end{table}

\paragraph{Results and Analyses} We present the results of zero-shot transfer experiments in~\cref{tab:scannet_transfer} and~\cref{tab:multiscan_transfer} with the following key observations: 
\begin{itemize}[nolistsep,noitemsep,label=$\centerdot$]
\item Our \ac{model} model demonstrates superior generalization to unseen scenes compared to the 3D-VisTA model. In zero-shot transfer scenarios, our model consistently outperforms 3D-VisTA across established benchmarks and \dataset-val. This indicates the effectiveness of contrastive alignment over traditional classification objectives, aligning with the advancements seen in \ac{2dvl} models for open-vocabulary grounding and transfer capabilities

\item \dataset dataset substantially enhances \ac{3dvl} grounding capabilities through zero-shot transfer, especially when provided with relatively limited training data, \ie, \dataset-val. As demonstrated in ~\cref{tab:multiscan_transfer}, there is a significantly improved performance when comparing models trained on \dataset in a zero-shot manner to those trained from scratch. This indicates that \dataset can effectively capture knowledge for general 3D scene grounding. Consequently, \textit{this underscores its potential as a go-to pre-training dataset for \ac{3dvl} tasks.}

\item The impact of our extensive collection and scalable generation of scene-text pairs is further evidenced by the results in the \textit{zero-shot text} setting. Notably, as shown in \cref{tab:scannet_transfer}, these automatically generated scene-text pairs supply ample knowledge for comprehending the scene distribution. This contributes significantly to the substantial improvement over the \textit{zero-shot} performance.

\end{itemize}



\begin{table}[t!]
    \begin{minipage}{\linewidth}
        \begin{minipage}[t]{0.47\linewidth}
            \centering
            \captionof{table}{\textbf{3D question answering results on ScanQA and SQA3D.} We report EM@1 score on ScanQA and SQA3D evaluation sets.}
            \vspace{-10pt}
            \resizebox{\linewidth}{!}{
                \begin{tabular}{ccccc}
                \toprule
                \multirow{2}[2]{*}{Model} & \multicolumn{3}{c}{ScanQA} & \multirow{2}[2]{*}{SQA3D} \\
                \cmidrule(lr){2-4}
                & val & w/obj & w/o obj & \\
                \midrule
                ScanRefer+MCAN~\cite{azuma2022scanqa} & 18.6 & 20.6 & 19.0 & - \\
                ScanQA~\cite{azuma2022scanqa} & 20.3 & 23.5 & 20.9 & 46.6 \\
                SQA3D~\cite{ma2022sqa3d} & - & - & - & 47.2\\
                3D-VisTA~\cite{zhu20233d} & 22.4 & \textbf{27.0} & 23.0 & 48.5\\
                3D-LLM~\cite{hong20233d}& 20.5 & 19.1 & - & -\\
                \midrule
                Ours & \textbf{22.7} & 25.0 & \textbf{23.5} & \textbf{49.9}\\
                \bottomrule
                \end{tabular}
            }
            \label{tab:3dqa}
        \end{minipage}
        \begin{minipage}[t]{0.505\linewidth}
            \captionof{table}{\textbf{Exisiting 3D backbones pre-trained on \dataset} for open-vocabulary 3D semantic segmentation on ScanNet. ``SPUNet'' denotes SparseUNet proposed in~\cite{yang2023regionplc}.}
            \centering
            \vspace{-10pt}
            \resizebox{\linewidth}{!}{
                \begin{tabular}{cccccc}
                    \toprule
                    Model & Network & mIoU & $\Delta$ & mAcc & $\Delta$\\
                    \midrule
                    OpenScene~\cite{peng2023openscene} & SPUNet16 & 57.2 & - & 69.9 & -\\
                    PLA~\cite{ding2023pla} & SPUNet16 & 17.7 & - & 33.5 & -\\
                    RegionPLC~\cite{yang2023regionplc} & SPUNet16 & 56.9 & - & 75.6 & - \\
                    RegionPLC+\dataset & SPUNet16 & \textbf{58.2} & +1.7\% & \textbf{77.3} & +2.2\% \\
                    \midrule
                    OpenScene~\cite{peng2023openscene} & SPUNet32 & 57.8 & - & 70.3 & - \\
                    PLA~\cite{ding2023pla} & SPUNet32 & 19.1 & - & 41.5 & - \\
                    RegionPLC~\cite{yang2023regionplc} & SPUNet32 & 59.6 & - & 77.5 & - \\
                    RegionPLC+\dataset & SPUNet32 & \textbf{61.0} & +2.3\% & \textbf{79.7} & +2.8\% \\
                    \bottomrule
                \end{tabular}
            }
            \label{tab:ov_seg}
        \end{minipage}
    \end{minipage}
    \vspace{-.2in}
\end{table}

\subsection{Additional \ac{3dvl} Tasks}\label{sec:exp:qa_seg}
\paragraph{Settings} We evaluate the effectiveness of \model and \dataset on additional \ac{3dvl} tasks: (i) \ac{3dqa} on ScanQA~\cite{azuma2022scanqa} and SQA3D~\cite{ma2022sqa3d}, and (ii) open-vocabulary 3D semantic segmentation (OV-Seg) on ScanNet.
\begin{itemize}[nolistsep,noitemsep,label=$\centerdot$]
\item In the \ac{3dqa} task, we follow Zhu \etal~\cite{zhu20233d} and evaluate models over the exact match metic (EM@1) on the validation and test sets of ScanQA, as well as the test set of SQA3D. We pre-train \model on \dataset and fine-tune the model on the \ac{3dqa} dataset to compare with \sota models.
\item In the \acs{ovseg} task, as \model builds upon an object-centric design and thus is not directly applicable to semantic segmentation, we consider testing the effectiveness of \dataset on improving existing 3D models. Specifically, we follow the open-vocabulary semantic segmentation settings proposed by Yang \etal~\cite{yang2023regionplc} and report the mIoU and mAcc score. We compare with existing works by pre-training the RegionPLC~\cite{yang2023regionplc} model on \dataset.
\end{itemize}

\paragraph{Results and Analyses} We present the results of \acs{3dqa} experiments in~\cref{tab:3dqa} and the results of \acs{ovseg} experiments in~\cref{tab:ov_seg}. The analyses are as follows:
\begin{itemize}[nolistsep,noitemsep,label=$\centerdot$]
\item As shown in~\cref{tab:3dqa}, our model achieves \sota results on both benchmarks, outperforming recent strong pre-training-based baselines like 3D-VisTA and 3D-LLM. As \dataset currently contains only descriptions of objects and scenes, we believe involving more types of language descriptions (\eg, question-answer pairs, dialogues) is a promising direction for further improving model performance on these downstream tasks.
\item As shown in~\cref{tab:ov_seg}, we observe consistent performance improvement of existing 3D backbone models on this task when pre-trained with \dataset data. This result validates that the collected data in \dataset can effectively boost the performance of existing models on scene understanding tasks. We further provide results of \sota 3D models that are pre-trained on \dataset on the close-vocabulary 3D semantic segmentation task in the \supp.
\end{itemize}

\subsection{Ablative Studies and Discussion}\label{sec:exp:ablation}
In this section, we present further discussions on both the data collected in \dataset and the \model model design. We aim to elucidate the effects of data scaling and show more clearly its effectiveness in 3D scene understanding. Regarding the experimental settings and more results discussion, refer to the \supp. The following points are specifically discussed in this section:

\paragraph{How important is data-scaling?} We conduct ablation studies over the amount of data used while pre-training \ac{model}. We consider the model trained with $\frac{1}{8}$, $\frac{1}{4}$, $\frac{1}{2}$ of \dataset to show the effectiveness of data-scaling on model performance in the pre-train and zero-shot transfer settings in ScanRefer and \dataset-val. As shown in~\cref{fig:data_scale}, we observe consistent performance improvement over the increase of data scale for both settings. We provide additional experiments in the \supp to show that such scaling effect is not only beneficial for \ac{3dvl} grounding but also for other 3D tasks like semantic segmentation~\cite{schult2023mask3d,yang2023swin3d}.

\begin{table}[t!]
    \begin{minipage}{\linewidth}
        \begin{minipage}{0.52\linewidth}
        \centering
        \includegraphics[width=0.49\linewidth]{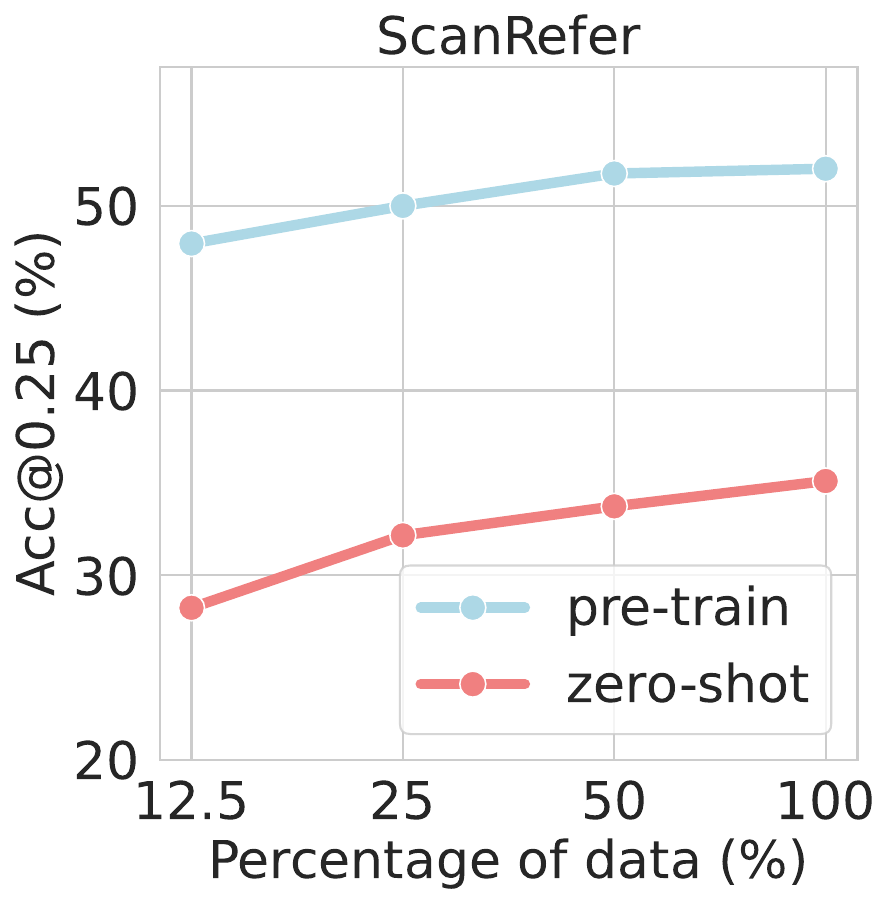}
        \hfill
        \includegraphics[width=0.49\linewidth]{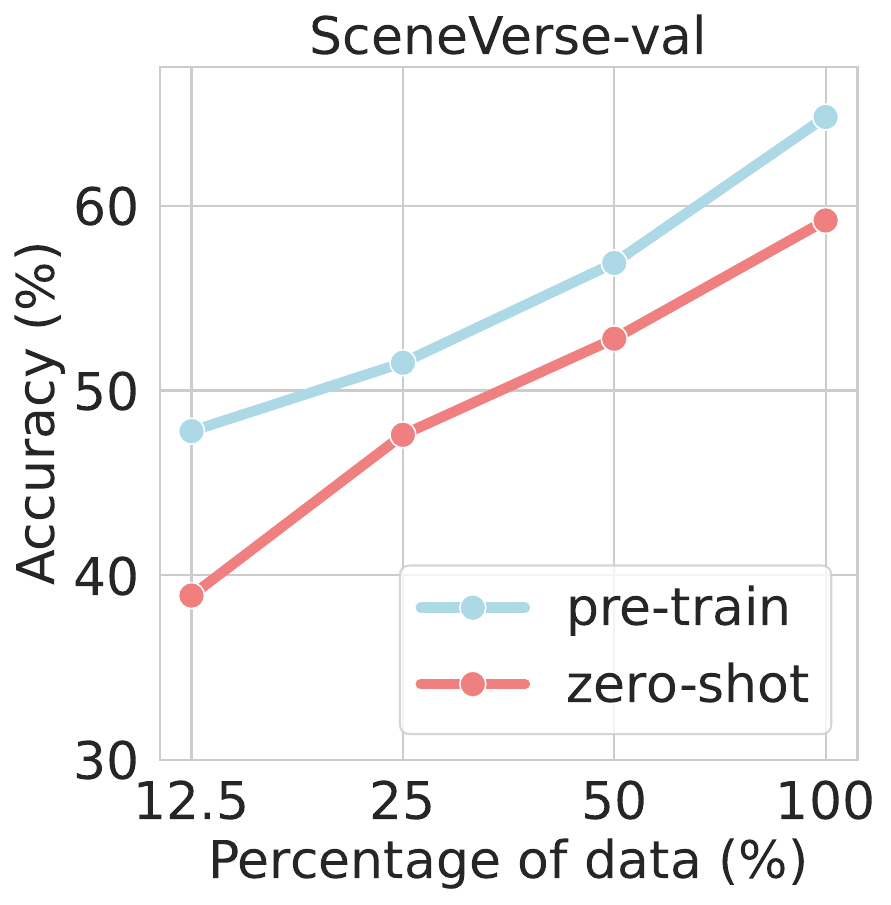}
        \captionof{figure}{\textbf{Model performance \textit{v.s.} data scale.} Plots show that models consistently improve in both the pre-train and zero-shot transfer settings on ScanRefer and \dataset-val with data scaling-up.}
        \label{fig:data_scale}
        \end{minipage}
        \hfill
        \begin{minipage}{0.46\linewidth}
            \centering
            \vspace{-.6in}
            \captionof{table}{\textbf{Ablation on text data source used in model pre-training}. All models are tested on ScanRefer with no additional finetuning.}
            \vspace{-10pt}
            \resizebox{\linewidth}{!}{
                \begin{tabular}{cccccc}
                \toprule
                Template & \acs{llm} & Anno. & Acc@0.25 & Acc@0.5\\
                \midrule
                \xmark & \xmark & \xmark & 43.5 & 38.4 \\
                \cmark & \xmark & \xmark & 50.9 & 46.1 \\
                \cmark & \cmark & \xmark & 51.1 & 46.3 \\
                \cmark & \cmark & \cmark & 52.0 & 47.1 \\
                \bottomrule
                \end{tabular}
            }
            \label{tab:data_ablation}
        \end{minipage}
    \end{minipage}
    \begin{minipage}{\linewidth}
        \begin{minipage}{0.52\linewidth}
            \centering
            \vspace{-.1in}
            \captionof{table}{\textbf{Cross domain transfer results} of models pre-trained on real and synthetic datasets. ``S3D'' stands for Structured3D.}
            \vspace{-10pt}
            \resizebox{\linewidth}{!}{
                \begin{tabular}{ccccc}
                \toprule
                Real & Synthetic & \dataset-val & S3D & ProcTHOR\\
                \midrule
                All & \xmark & 64.8 & 37.1 & 43.4 \\
                \xmark & S3D & 7.0 & 85.1 & 16.1 \\
                \xmark & ProcTHOR & 4.2  & 16.3 & 91.0 \\
                \bottomrule
                \end{tabular}
            }
            \label{tab:synthetic_dataset}
        \end{minipage}
        \hfill
        \begin{minipage}{0.46\linewidth}
            \vspace{-.55in}
            \captionof{table}{\textbf{Ablation on model design on \dataset-val.} We use ``Obj-lvl'', ``Scene-lvl'' to denote object and scene alignment loss, and ``MLM'' for the mask language modeling loss.}
            \centering
            \vspace{-10pt}
            \resizebox{\linewidth}{!}{
                \begin{tabular}{cccccccc}
                \toprule
                Obj-lvl. & MLM & Scene-lvl. & Overall & Easy & Hard \\ 
                \midrule
                \xmark & \xmark & \xmark & 64.8 & 75.4 & 48.7 \\ 
                \cmark & \xmark & \xmark & 65.2 & 77.1 & 47.4 \\ 
                \cmark & \cmark & \xmark & 62.4 & 73.4 & 45.8 \\ 
                \cmark & \cmark & \cmark & \textbf{66.9} & \textbf{77.8} & \textbf{50.3} \\ 
                \bottomrule
                \end{tabular}
            }
            \label{tab:model_ablation}
        \end{minipage}
    \end{minipage}
    \vspace{-.1in}
\end{table}

\paragraph{How is the generated data compared with human-annotated data?}  We assess the performance of models trained using various scene-text sources, specifically focusing on their performance in the ScanRefer dataset without additional fine-tuning. As shown in~\cref{tab:data_ablation}, models trained with our template-based generated texts and \ac{llm}-refined texts show significant improvements over models trained solely on ScanRefer. More importantly, these variants of our model already achieve \sota results compared with previous baselines. This indicates the effectiveness of our text-generation pipeline. Finally, we observe that adding human-annotated data is still beneficial for model performance. However, the improvement is relatively marginal over models trained on our generated data. 

\paragraph{What is the role of the synthetic scenes in this scale-up process?} With synthetic data providing large-scale and diverse scene data for \ac{3dvl} tasks, we evaluate the models' domain transfer (Sim2Real) capability. Specifically, we compare models trained on all real scenes in \dataset against models trained exclusively on two synthetic subsets of \dataset, \ie, Structured3D and ProcTHOR. As shown in~\cref{tab:synthetic_dataset}, models trained on synthetic subsets demonstrate remarkable performance on their corresponding test sets while suffering when transferred to real or other synthetic scenes. In contrast, the model trained on real scene-text pairs exhibits less performance drop when generalizing to synthetic scenes. This result affirms the domain gap between real and synthetic scenes in \ac{3dvl} grounding and shows that a simple scale-up in the number of scenes is insufficient when naturalness can not be guaranteed. Considering the scalability of our language generation pipeline and the scaling effect shown in our experiments, the rate-determining step for further scaling-up \ac{3dvl} comes to the collection of diverse, high-quality, and realistic scenes that capture natural 3D scene distributions.

\paragraph{How important is the design of each module in \model?} We provide ablative analyses of our multi-level contrastive alignment design in~\cref{tab:model_ablation}. We mainly consider removing objectives in our model to reveal the effectiveness of each level of alignment. We choose the referral-object-level alignment objective as the default setting and consider removing: (i) object-level alignment objective, (ii) masked language modeling objective, and (iii) scene-level alignment objective. 
When removing the object-level alignment objective, we learn the object point cloud encoder with the referral-object-level alignment and without pre-training. 
As shown in~\cref{tab:model_ablation}, we test different models on the \dataset-val without additional fine-tuning. Results show that the scene-level alignment objective is crucial for referral object grounding in \dataset-val with the $\sim$5\% performance drop. Similar observations could be made for the model trained without object-level alignment ($\sim$2\% drop) and masked language modeling objective ($\sim$1.5\% drop). These results affirm the effectiveness of our model design.

\section{Conclusion}\label{sec:conclusion}
In this work, we scale up \ac{3dvl} for grounded scene understanding. We present \dataset, a million-scale \ac{3dvl} dataset covering various scenes and multi-level scene descriptions sourced from both human annotation and our proposed scene-text generation approach. Utilizing \dataset, we propose Grounded Pre-training for Scenes (GPS), a model trained with multi-level scene-language contrastive alignment. Through extensive experiments, we show that \model achieves \sota results on common \ac{3dvl} tasks including grounding and question answering. We further conduct zero-shot transfer experiments to show the improved generalization performances of \model trained on \dataset compared with previous baselines. We also demonstrate that the scaling effect of \dataset is generally beneficial for existing 3D models on \ac{3dvl} tasks like semantic segmentation. We hope our efforts and successful scale-up attempts in \dataset could pave the way for new research paradigms in \ac{3dvl}.

\paragraph{Acknowledgement} The authors would like to thank Yaowei Zhang (BIGAI) for his help on online visualization and other colleagues from BIGAI General Vision Lab for fruitful discussions. The authors would also like to thank the anonymous reviewers for their constructive feedback. 

\bibliographystyle{splncs04}
\bibliography{reference}

\clearpage
\appendix
\setcounter{page}{1}
\renewcommand{\thefigure}{A.\arabic{figure}}
\renewcommand{\thetable}{A.\arabic{table}}
\renewcommand{\theequation}{A.\arabic{equation}}
\setcounter{section}{0}
\setcounter{figure}{0}
\setcounter{table}{0}
\setcounter{equation}{0}
{
    \Large
    \begin{center}
    \textbf{\textsc{SceneVerse}:} \textbf{Scaling 3D Vision-Language Learning for Grounded 
    Scene Understanding} \\
    \vspace{0.5em}
    Supplementary Material
    \vspace{0.5em}
    \end{center}
}


\section{The \texorpdfstring{\dataset}{} Dataset}\label{supp:dataset}

\subsection{3D Scenes}
To address the scarcity of available 3D scene data, we construct \dataset by unifying 3D scene data from various existing datasets. The curation involves utilizing real-world scene datasets such as \scannet~\cite{dai2017scannet}, \arkitscene~\cite{baruch2021arkitscenes}, \hmthreeD~\cite{ramakrishnan2021habitat}, \rscan~\cite{wald2019rio} and \multiscan~\cite{mao2022multiscan},in conjunction with synthetic environments from \structured~\cite{zheng2020structured3d} and \procthor~\cite{deitke2022️}. The incorporation of these synthetic datasets is primarily driven by their potential as scalable data sources for \ac{3dvl} alignment.
To facilitate the training process, we conduct the following preprocessing steps.
\paragraph{Room Segmentation} The 3D scenes in \hmthreeD and \procthor are released at the building level, encompassing  multiple rooms and sometimes spanning over 50 meters. To align with existing benchmarks~\cite{chen2020scanrefer,achlioptas2020referit3d}, we leverage the associated metadata to segment the 3D point cloud at the room level, facilitating subsequent operations in scene graph construction and language description generation. Additionally, we implement a filtering process to exclude extremely large rooms and those with fewer than 4 objects in the scene.
\paragraph{Point Cloud Normalization} To mitigate the data disparities arising from diverse capture devices across various data sources, we subsample each point cloud to a maximum of $240,000$ points. Each point cloud then undergoes a transformation centered on the central point on the floor, followed by rotation to align the room layout with the axis following the approach by Chen \etal~\cite{chen2022language}.

\paragraph{Semantic Label Alignment} Given the divergence in semantic label sets across different datasets, we undertake a comprehensive effort to map all the object class labels to the $607$ semantic labels in ScanNet~\cite{dai2017scannet} to facilitate close-vocabulary object classification~\cite{qi2017pointnet++} in the existing model framework~\cite{zhu20233d}. We construct the mapping in each dataset through LLM and manual verification. Note that the object-level grounding in \model can directly deal with open-set object labels or captions, similar to CLIP~\cite{hegde2023clip}.

After the preprocessing, each scan is represented by a point cloud $\mathrm{P} \in \mathbb{R}^{N\times8}$, wherein each point is defined by its 3D coordinates, RGB color, instance id and semantic label. In total, we curate \sceneNumTotal 3D scenes in \dataset.

\subsection{3D Scene Graph Construction}
In \cref{sec:data_scene_graph}, we introduce an automated pipeline to construct 3D scene graphs from point clouds. Here, we provide more implementation details and the relationship definition.

\begin{figure}[t!]
    \centering
    \includegraphics[width=0.9\linewidth]{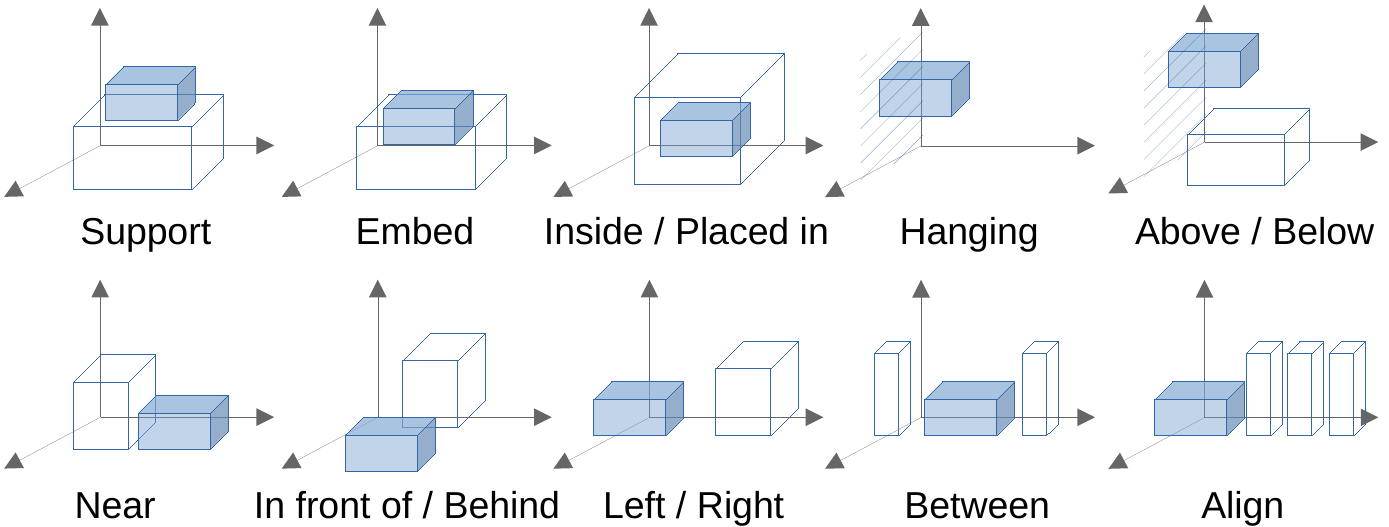}
    \caption{\textbf{Overview of the relationships in \dataset.} The target object is colored in \textcolor{blue}{blue}.}
    \label{fig:relations}
\end{figure}

\paragraph{Relationships} 
\begin{table}[t]
    \caption{\textbf{Relationships in \dataset.} The 3D scene graph captures 21 types of relationships ranging in 4 categories.}
    \centering
    \resizebox{0.7\linewidth}{!}{
        \begin{tabular}{ccc}
        \toprule
        \multicolumn{1}{c}{Category} & \multicolumn{2}{c}{Relation}\\
        \midrule
        \multirow{2}{*}{In-contact vertical}  & supported by  &  embedded into \\
        & placed in & inside \\
        \midrule
        \multirow{4}{*}{Non-contact vertical}  & hanging on & affixed on \\
         & mounted on & above \\
        & higher than & below  \\
         & lower than \\
        \midrule
        \multirow{4}{*}{Horizontal}  & near(far) to the left of & near(far) to the right of \\
         & is behind & is in front of \\
         & close to & adjacent to \\
         & besides & next to \\
        \midrule
        Multi-object  & between & aligned \\
        \bottomrule
        \end{tabular}
    }
    \label{tab:relations_type}
\end{table}
Our 3D scene graph captures \NumOfRel types of relations as shown in Tab.~\ref{tab:relations_type}. 
We provide illustrations of how these relations are defined in the 3D space, as can be seen in ~\cref{fig:relations}.

\paragraph{Scene Graph Construction} 
Due to the inherent noise and incompleteness in the point cloud representation, automatically extracting precise and comprehensive relationships from the point clouds is a non-trivial task. Below we detail our 3D scene graph construction process, as outlined in~\cref{supp:alg:pseudo_code_ssg_construct}. 

We first instantiate the graph nodes with the instance annotation from the point cloud and parameterize each node with object centroid $\mathrm{p_i} \in \mathbb{R}^3$ and size of the axis-aligned bounding box $\mathrm{b_i}=(b_x,b_y,b_z) \in \mathbb{R}^3$ (Line 1-3). Next, we traverse all the nodes to determine their spatial relationships (Line 4-22). Notably, in cases where an object node lacks any in-contact vertical relationships with other objects in the scene, we designate such objects as \textit{"hangable"} and calculate their non-contact vertical relationships (Line 9-13). Examples of such objects include paintings, curtains, \etc. Finally, we establish relationships between multiple objects (Line 23): i) When a target object is connected with two edges labeled \texttt{left} and \texttt{right}, the target object, along with the two neighboring nodes, forms a \texttt{between} relationship triplets. ii) If the offset of the center point coordinates of a group of objects in either the X-axis or Y-axis direction is smaller than a specified offset threshold $\delta$, then this group of objects forms an \texttt{align} relationship. The offset threshold $\delta$ will be adjusted based on the size of the scene. 
In additional, we utilize an automatic verification procedure to validate the scene graph, further improving the quality of the scene graph we constructed (line 24). One of the verification operations involves manually maintaining a mapping between objects and relationship descriptions based on common sense. For example, people usually use ``mounted on'' to describe the relation between TV and wall, rather than ``hanging on''. Therefore, we would automatically 
 refined (~\text{\textcolor{ForestGreen}{TV}}, \text{\textcolor{blue}{hanging on}}, ~\text{\textcolor{YellowOrange}{wall}}) to (~\text{\textcolor{ForestGreen}{TV}}, \text{\textcolor{blue}{mounted on}}, ~\text{\textcolor{YellowOrange}{wall}}).

In our constructed 3D scene graph $\mathcal{G}=(\mathcal{V},\mathcal{E})$, the nodes $\mathcal{V}$ comprises the union of node sets $\mathcal{V}_1\bigcup\mathcal{V}_2\bigcup\dots\bigcup\mathcal{V}_K$, with $\mathcal{V}_k$ representing the set of nodes at a particular hierarchical level. The hierarchies are determined by the \texttt{support} relationship; for instance, objects supported by the floor constitute $\mathcal{V}_0$, while objects supported by the table will form $\mathcal{V}_1$, \etc. Note that edges originating from one node $v\in \mathcal{V}_k$ may only terminate in nearby hierarchies $\mathcal{V}_k\cup \mathcal{V}_{k+1}\cup \mathcal{V}_{k+1}$. In other words, edges in the scene graph exclusively connect nodes within the same hierarchical level, or one level higher or lower.

\begin{algorithm}
\caption{Scene Graph Construction Pipeline}
    \SetKwInOut{Input}{Input}
    \SetKwInOut{Output}{Output}
    \Input{ 
        $M$ object point clouds $\{P_{1}, P_{2}, \dots, P_{m}\}$
    }
    \Output{3D scene graph $\mathcal{G}(\mathcal{V},\mathcal{E})$}
    \begin{algorithmic}[1]
    \FOR{$i$ from $1$ to $M$}
        \STATE Create node $v_{i} \in \mathcal{V}$ using the centroid $p_{i}$ and bounding box size $b_{i}$ of object\\ point cloud $P_{i}$ 
        \ENDFOR 
        \FOR{$i$ from $1$ to $M$}
            \FOR{$j$ from $i+1$ to $M$}
            \STATE $\mathrm{RelsType}_{v} \gets \mathrm{VerticalInContact}(v_{i},v_{j})$
            \STATE Add in-contact vertical relationship triplets $(v_{i}, v_{j}, e_{i,j})$ with $\mathrm{RelsType}_{v}$ \\to $\mathcal{G}$
            \ENDFOR
        \IF {No objects horizontally related to $v_{i}$}
            \FOR{$k$ from $1$ to $M$ and $i \neq k$}
                \STATE $\mathrm{RelsType}_{v} \gets \mathrm{VerticalNonContact}(v_{i},v_{k})$
                \STATE Add non-contact vertical relationship triplets $(v_{i}, v_{k}, e_{i,k})$ with \\ $\mathrm{RelsType}_{v}$ to $\mathcal{G}$
            \ENDFOR
        \ENDIF
    \ENDFOR
    \FOR{ $v_{i} \in  \mathcal{V} $}
        \STATE let ${\{v_{i_1}, v_{i_2},...,v_{i_N}\}}$ be the $N$ different nodes with the same in-contact \\ vertical parent node $v_{i}$
        \FOR {$j$ from $1$ to $N$}
            \STATE $\mathrm{RelsType}_{h} \gets \mathrm{Horizontal}(v_{i},v_{i_j})$
            \STATE Add horizontal relationship triplets $(v_{i}, v_{i_j}, e_{i,i_j})$ with $\mathrm{RelsType}_{h}$ to $\mathcal{G}$
        \ENDFOR
    \ENDFOR
    \STATE Update $\mathcal{G} \gets \mathrm{MultiObjects}(\mathcal{G})$
    \STATE Update $\mathcal{G}$ with  automatic verification procedure
    \end{algorithmic}

\label{supp:alg:pseudo_code_ssg_construct}
\end{algorithm}





\subsection{Language Generation Details}
In \cref{sec:data_language}, we adopt both templates and \ac{llm} to automatically generate scene-language pairs in \dataset. More technical details and examples are provided in this section.

\subsubsection{Object Captioning Pipeline}
\label{supp:object_caption}
Object captions aim to provide detailed descriptions of an object's visual and physical properties, facilitating object-level grounding with its distinctive features. The detailed object captioning pipeline is outlined in \cref{supp:alg:pseudo_code_object_caption}. Given the multi-view images $\{I_1, I_2, \dots, I_n\}$, we utilize the point cloud $P_o$ of the object $o$ to get the visible points $P_{o,v}^{vis}$ in the images $v$ through rendering. The occlusion score $s_{o,v}^{occ}$ is calculated as the ratio between the number of visible points and the object point cloud. The image is then cropped with the rendered bounding box and processed through BLIP2~\cite{li2023blip2} to generate the initial object caption $C_{o,v}$. For each initial caption, we calculate its CLIP~\cite{radford2021learning} similarity score between the text and the cropped image, denoted by $s_{o,v}^{clip}$. To get a refined object caption, we select the top $10$ initial captions with the highest CLIP score and minimal occlusion. The selected sentences are fed into a \acs{llm} to obtain a coherent summary of the object captions. In this process, we explicitly instruct the language model to identify and correct the potential errors.

\begin{figure*}[ht!]
    \centering
    \includegraphics[width=0.9\linewidth]{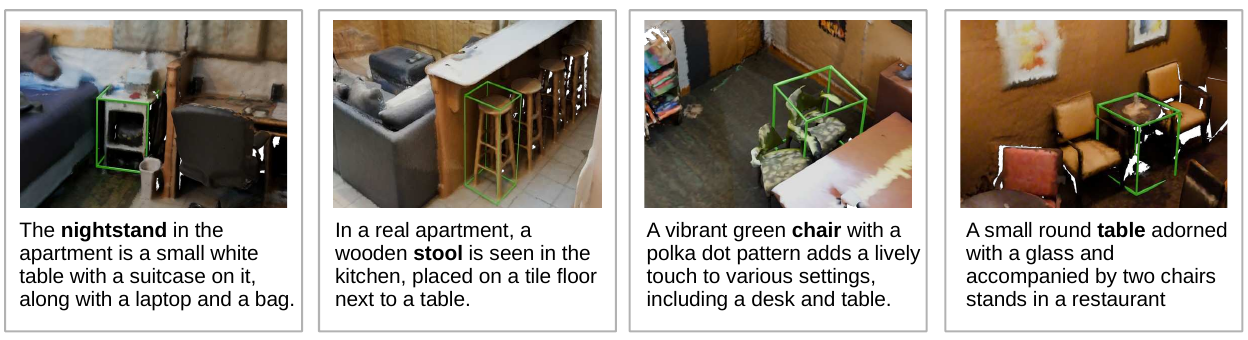}
    \caption{\textbf{Examples of object captioning.} We color the target object in \textbf{bold}.}
    \label{fig:obj_cap}
\end{figure*}%
\begin{figure*}[t!]
    \centering
    \includegraphics[width=0.9\linewidth]{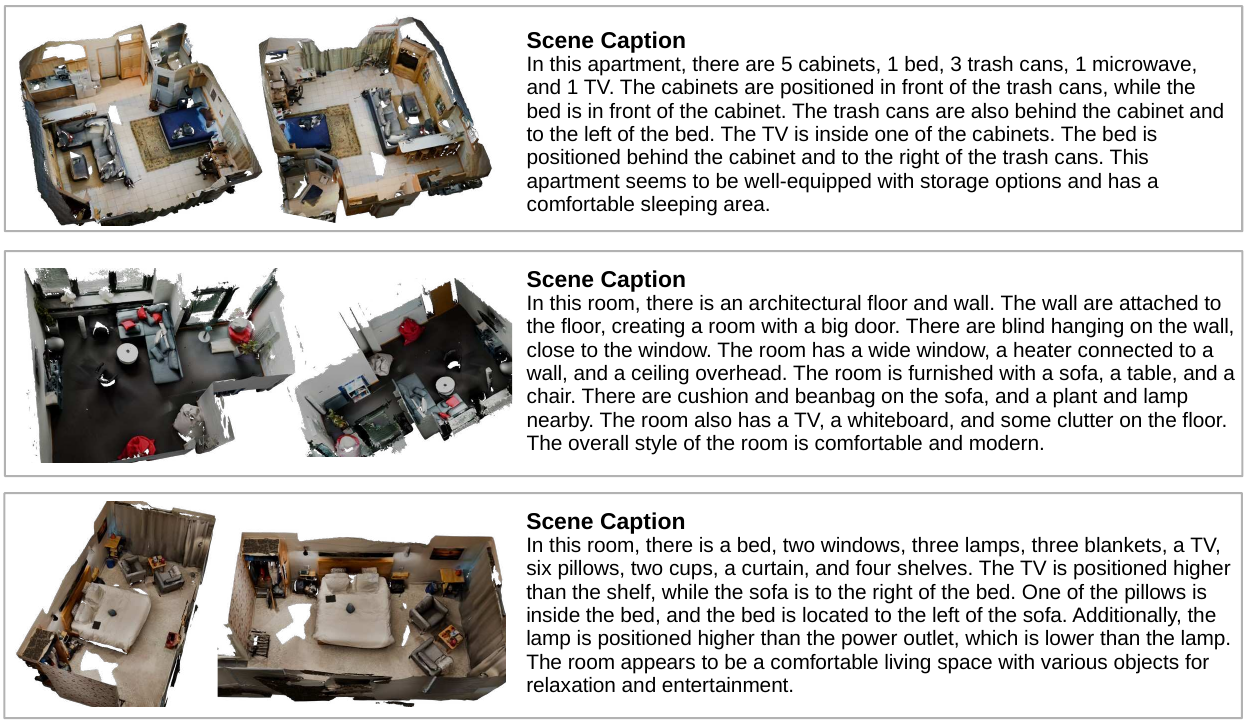}
    \caption{\textbf{Examples of scene captioning.}}
    \label{fig:scene_cap}
\end{figure*}%
\begin{figure*}[t!]
    \centering
    \includegraphics[width=0.9\linewidth]{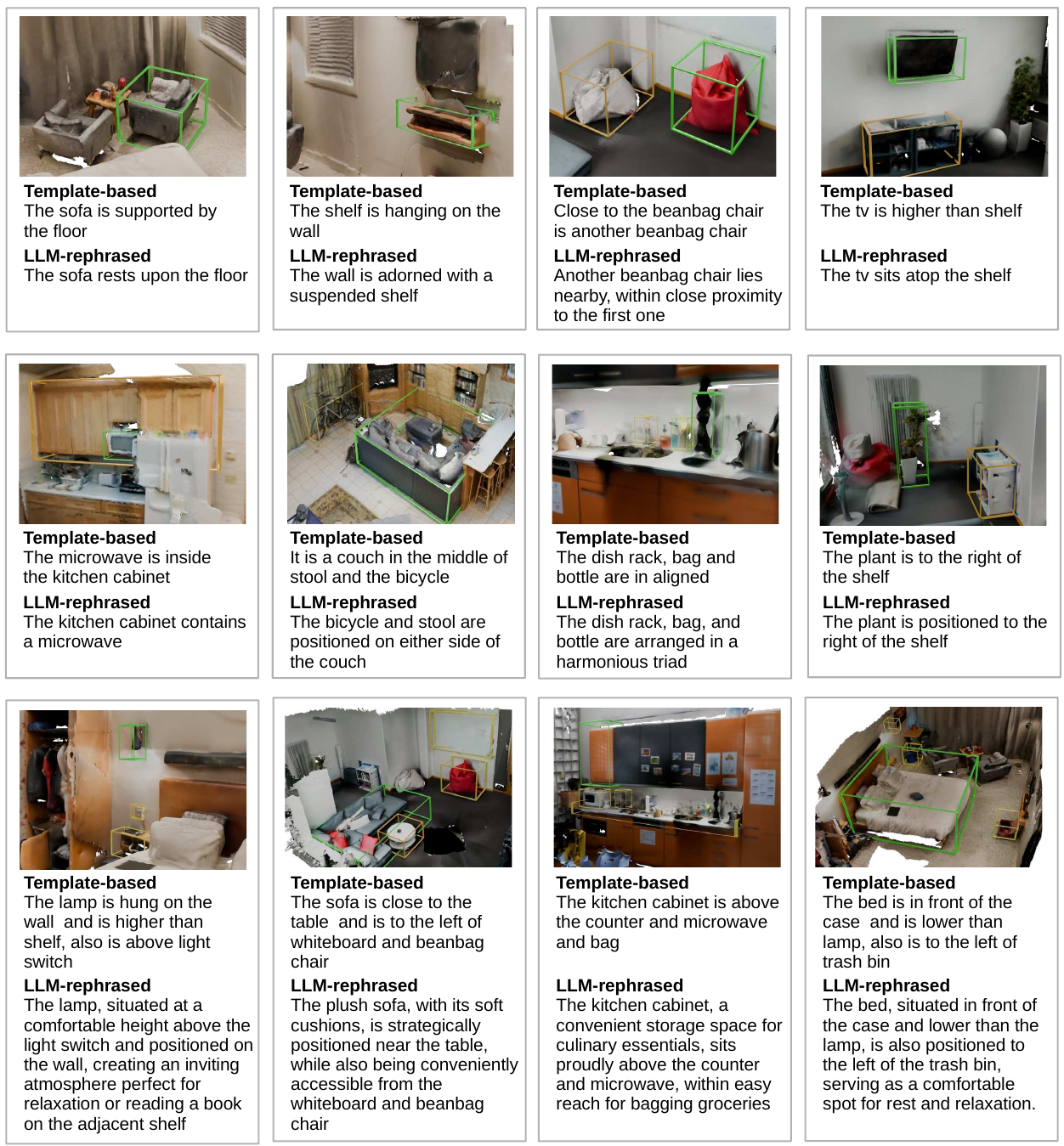}
    \caption{\textbf{Examples of object referral.} Note that  \textcolor{ForestGreen}{the green bounding box} indicates the target object and \textcolor{YellowOrange}{yellow bounding box} indicates the anchor object(s).}
    \label{fig:obj_ref}
\end{figure*}%

\begin{algorithm}
\caption{Object Captioning Pipeline}
\SetKwInOut{Input}{Input}
\SetKwInOut{Output}{Output}
\Input{ 
    $M$ object point clouds $\{P_{1}, P_{2}, \dots, P_{m}\}$;
    $N$ multiview images $\{I_{1}, I_{2}, \dots, I_{n}\}$
}
\Output{Captions for each object in the scene $\{C_{1}, C_{2}, \dots, C_{m}\}$}
\begin{algorithmic}[1]
\FOR{$o=1,2,\dots,M$}
    \FOR{$v=1,2,\ldots,N$}
    \STATE Project $P_{o}$ on $I_{v}$ to get visible points $P^{vis}_{o,v}$ 
    \STATE Crop $I_{v}$ with the bounding box of $P^{vis}_{o,v}$ to get $I^{crop}_{o,v}$
    \STATE Get the image caption $C_{o,v}$ for $I^{crop}_{o,v}$ using BLIP2~\cite{li2023blip2}
    \STATE Calculate the similarity score $s^{clip}_{o,v}$ between $C_{o,v}$ and $I^{crop}_{o,v}$ with \\ CLIP~\cite{radford2021learning}
    \STATE Calculate the occlusion score $s^{occ}_{o,v} = \frac{\#P^{vis}_{o,v}}{\#P_{o}}$
    \ENDFOR
\STATE Select the top-10 $\{C_{o,v}\}$ with highest $s^{clip}_{o,v}*s^{occ}_{o,v}$ 
\STATE Summary selected $\{C_{o,v}\}$ with GPT-3.5 to get $C_{o}$
\ENDFOR
\end{algorithmic}
\label{supp:alg:pseudo_code_object_caption}
\end{algorithm}

\subsubsection{Automatic Language Generation}
\begin{enumerate}
\item \textbf{Template-based} We create diverse templates to generate descriptions for each type of relationship. We categorized the templates into three types based on the number of objects involved and their spatial relationships.
\begin{itemize}[leftmargin=2em]
    \item \textbf{Pair-wise}: The pair-wise templates are used to describe the positional relationship between the target object and the anchor object in the scene. We design various templates to enrich the templated-based descriptions, spanning active and passive tense, as well as inversion clauses. Typical examples are shown below:
    \begin{itemize}[leftmargin=1em] \small
    
    \item[-]  The~\text{\textcolor{ForestGreen}{target-object}}~(is)~\text{\textcolor{blue}{spatial-relation}}~the~\text{\textcolor{YellowOrange}{anchor-object}}.

    \item[-] It is a~\text{\textcolor{ForestGreen}{target-object}} that (is)~{\textcolor{blue}{spatial-relation}}~the~{\textcolor{YellowOrange}{anchor-object}}.

    \item[-] There is a~{\textcolor{ForestGreen}{target-object}} that (is)~{\textcolor{blue}{spatial-relation}}~the~{\textcolor{YellowOrange}{anchor-object}}.

    \item[-] {\textcolor{blue}{Spatial-relation}}~the~{\textcolor{YellowOrange}{anchor-object}} is the~{\textcolor{ForestGreen}{target-object}}.

    \item[-] {\textcolor{blue}{Spatial-relation}}~the~{\textcolor{YellowOrange}{anchor-object}}, a~{\textcolor{ForestGreen}{target-object}} is placed.

    \end{itemize}
    
    \item \textbf{Multi-objects}: This is utilized when the target object forms a \texttt{between} or \texttt{align} relationship with multiple anchor objects in the scene. The templates follow the same construction rules as the \textbf{Pair-wise} templates.
    \item \textbf{Star-reference}: To increase complexity in templated-based descriptions, we design ``star-reference'' to describe the target object and its relationship with 3 randomly selected anchor objects in the scene graph. In particular, we perform cluster analysis on the selected relationship triplets. Based on the diversity of the analysis, different templates will be chosen to generate descriptions. 
    For example, when the relations between 3 anchor objects and the target object is the same, we prefer to use the template like: ``The~{\textcolor{ForestGreen}{target-object}}~(is)~{\textcolor{blue}{spatial-relation}}~the~{\textcolor{YellowOrange}{anchor-object-1}},~{\textcolor{YellowOrange}{anchor-object-2}} and~{\textcolor{YellowOrange}{anchor-object-3}}''. If 2 out of the 3 anchor objects have the same relations with the target object, we would use a template like: ``The~{\textcolor{ForestGreen}{target-object}}~(is)~{\textcolor{blue}{spatial-relation-1}}~the~{\textcolor{YellowOrange}{anchor-object-1}} and~{\textcolor{YellowOrange}{anchor-object-2}}, and (is)~{\textcolor{blue}{spatial-relation-2}}~the~{\textcolor{YellowOrange}{anchor-object-3}}.''

\end{itemize}

\begin{table*}[ht]
    \caption{\textbf{Prompts used in \dataset.}}
    \centering
    \resizebox{\linewidth}{!}{
        \begin{tabular}{cp{15cm}}
        \toprule
        Description type & Prompt \\
        \midrule
        Object caption & Summarize \textcolor{objcaption}{\textbf{caption}} below. The summary should be a description of the~\text{\textcolor{ForestGreen}{\textbf{target-object}}}. Focus on the~\text{\textcolor{ForestGreen}{\textbf{target-object}}}'s attribute, like color, shape and material, \etc. Identify and correct the potential errors. 
        \newline \textcolor{objcaption}{\textbf{caption}}: \textit{A bed in a hotel room. A white comforter on a bed. A bed with a striped comforter...}
        \newline \text{\textcolor{ForestGreen}{\textbf{target-object}}}: \textit{Bed}
        \\

        \midrule
        Object referral & \begin{tabular}[t]{p{15cm}}
        Rewrite the following~\textcolor{objrefer}{\textbf{caption}} using one random sentence structure. You should give me only one rewritten sentence without explanation.\\
        \textcolor{objrefer}{\textbf{caption}}: \textit{The bed is between desk and nightstand.}
        \end{tabular} \\
        
        \cline{2-2}
        & \begin{tabular}[t]{p{15cm}}
        
        Rewrite the following~\textcolor{objrefer}{\textbf{caption}}. You should give me only one rewritten sentence about~\text{\textcolor{ForestGreen}{\textbf{target-object}}} without explanation. Make sure~\text{\textcolor{ForestGreen}{\textbf{target-object}}} is the subject of the sentence, not~\text{\textcolor{YellowOrange}{\textbf{anchor-object(s)}}}. If the sentence is in full inversion, keep the inversion. \\
        \textcolor{objrefer}{\textbf{caption}}: \textit{The armchair is next to the sofa.} \\
        \text{\textcolor{ForestGreen}{\textbf{target-object}}}: \textit{Armchair} \\
        \text{\textcolor{YellowOrange}{\textbf{anchor-object(s)}}}: \textit{Sofa}
        \end{tabular} \\
        
        \cline{2-2}
        & \begin{tabular}[t]{p{15cm}}
        Rewrite the following~\textcolor{objrefer}{\textbf{caption}} using one random sentence structure. You need to focus on the location and relations of the~\text{\textcolor{ForestGreen}{\textbf{target-object}}} that appears in the sentence. If multiple~\text{\textcolor{ForestGreen}{\textbf{target-object}}} appear in the sentence, you need to focus on the first~\text{\textcolor{ForestGreen}{\textbf{target-object}}} that appears. You can also add the~\text{\textcolor{ForestGreen}{\textbf{target-object}}}'s function and comfort level based on the sentence, e.g., how the objects can be used by humans and human activities in the scene. You should give me only one rewritten sentence without explanation. \\
        \textcolor{objrefer}{\textbf{caption}}: \textit{Far from the bowl and peppershaker, the vase is to the left, it is also on the top of countertop.} \\
        \textcolor{ForestGreen}{\textbf{target-object}}: \textit{Vase}
        \end{tabular} \\

        \midrule
        Scene captioning & Your task is to provide a summary for a scene from a given \textcolor{scenecaption}{\textbf{scene graph}}. The scene contains some objects, which compose a scene graph in json format. 
        \newline There are 3 types of descriptions in scene graph: ``scene type'' denotes the type of the scene. ``objects count'' then listed the objects in the scene and their quantity, it should be noted that the actual objects in the room may be more than listed. ``objects relations'' describe the spatial relations with objects. 
        \newline Also describe the scene concerning commonsense, e.g., how the objects can be used by human and human activity in the scene. The description should conform to the given scene graph. The spatial relations between objects can only be inferred from the ``objects relations`` in scene graph. Don't describe each object in the scene, pick some objects of the scene for summary. Don't describe each relations in the scene, pick some relations of the scene for summary. You can also summarize the room's function, style, and comfort level based on the arrangement and count of objects within the room. The summary should be about the object types, object attributes, relative positions between objects. Your summary must not exceed 80 words. You must write using one random sentence structure. 
        \newline \textcolor{scenecaption}{\textbf{scene graph}}: {\ttfamily
        \{ `scene\_type': `Bedroom', `object\_count': \{`nightstand':2, ...\}, `relation': \{`nightstand', `on', `floor'\}, \{`backback', `in front of', `bed'\}, ...\}
        }\\
        \bottomrule
        \end{tabular}
    }
    \label{tab:prompt_type}

\end{table*}

\item \textbf{\ac{llm}-rephrasing} To increase description diversity we use the GPT-3.5\cite{openai2022chatgpt}~and Llama\cite{touvron2023llama}~for description rephrasing. This improves the diversity and naturalness of the template-based descriptions, as is shown in \cref{fig:stats}. The detailed prompts are provided in Tab.~\ref{tab:prompt_type}.
\end{enumerate}

More examples of the scene-language pairs in \dataset are shown in \cref{fig:obj_cap}, \cref{fig:scene_cap} and \cref{fig:obj_ref}.

\section{Model Details}\label{sup:model}
\subsection{Spatial-Attention Transformer}\label{sup:model:sattn}
In~\cref{sec:model:scene}, we leveraged and spatial-attention based transformer architecture to aggregate object-level point cloud features with spatial location information. In this section, we provide the detailed design of this proposed module. 

Formally, given object features $\{\vf_i^{O}\}_{i=1}^{N}$ and their locations $\{\vl_i\}_{i=1}^{N}$, we first construct pair-wise spatial relationship feature via:
\begin{equation*}
    \vm_{ij} = \left[d_{ij}, \sin(\theta_h), \cos(\theta_h), \sin(\theta_v), \cos(\theta_v)\right],
\end{equation*}
where $d_{ij}$ denotes the Euclidean distance between objects and $\theta_h, \theta_v$ are the horizontal and vertical angles of the line connecting the centers of objects $i$, $j$. We then use this pair-wise relationship feature $M = [\vm_{ij}] \in \mathbb{R}^{N\times N \times 5}$ to modulate the attention weights of the self-attention layer in the transformer when aggregating object features as shown below:
\begin{equation*}
    \small
    \mathrm{Attn}(Q,K,V,M) = \mathrm{softmax}\left(\frac{QK^T}{\sqrt{d_h}} + \log\sigma(M\omega)\right)V,
\end{equation*}
where $\omega \in \mathbb{R}^{5}$ is a projection layer mapping spatial pair-wise features to the attention scores and $\sigma$ denotes the sigmoid function. This process could be equivalently interpreted as using the spatial location of objects to adjust the self-attention feature aggregation between objects, making spatially related objects have more attention weights.

\subsection{Pre-training Details}\label{sup:model:implement}
For training our model \model, we conduct a two-stage training approach. As described in~\cref{sec:model:refer}, we first pre-train the object point cloud encoder with the object-level grounding objective. Next, we freeze the object point cloud encoder during the second pre-training stage for scene-level pre-training that includes model training with scene-level grounding and referral object grounding objectives. This design is inspired by recent works like~\cite{zhu20233d,chen2023unit3d} that demonstrated a well-grounded initialization of object representations is beneficial for 3D scene grounding.

\paragraph{Object-level pre-training} To correctly align objects in scenes with their captions, we utilize the ground-truth object bounding boxes provided with the datasets to segment all objects in the scenes. Next, we utilize a PointNet++~\cite{qi2017pointnet++} encoder to encode and align these object point clouds with object captions provided in \dataset following~\cref{sec:model:object}. For object instances with no object captions synthesized, we follow~\cite{radford2021learning} and construct captions with their semantic class labels like ``the point cloud of <\texttt{CLASS}>''. Notably, as our model design sets no constraints on object point cloud encoders, the choice of object encoder mainly depends on the computing resources available.

\paragraph{Scene-level pre-training} With pre-trained object feature extractors, we further use both scene captions and object-referring expressions for scene-level pre-training. We use a 4-layer BERT encoder for encoding both scene captions and object referrals. As discussed in~\cref{sec:model:scene}, we apply a 4-layer spatial transformer to encode object features with their locations. For scene-level grounding, we adopt a max-pooling layer to aggregate features from the spatial transformer and align with the \texttt{[CLS]} feature of the scene caption. For referral-object-level grounding, we further pass the obtained object features as well as the referral language features into a 4-layer self-attention transformer and use the grounding objective described in~\cref{sec:model:refer} to match the referred object's feature and the \texttt{[CLS]} feature of the referring expression.

\paragraph{Training} For object-level pre-training, we utilize an AdamW optimizer with a learning rate of $1\times10^{-2}$ for 1500 epochs and no warm-up periods. During training, we use a batch size of 512 and leverage a cosine annealing scheme for learning rate scheduling with a minimum learning rate of $1\times10^{-3}$. For scene-level pre-training, we use an AdamW optimizer with a learning rate of $1\times10^{-5}$ for the language encoder, a learning rate of $1\times10^{-4}$ for the spatial transformer, a learning rate of $1\times10^{-4}$ for the self-attention transformer, and a learning rate of $5\times10^{-4}$ for all remaining learnable parameters (\eg, projections). For all experiments, we train the model for 150 epochs with a warm-up period of 500 and also a cosine annealing scheme for learning rate with a minimum learning rate ratio of $0.1$. All pre-training experiments are run on 8 NVIDIA-A100 GPUs with the longest pre-training on \dataset taking about 2 days.

\section{Experimental Details}\label{supp:exp}
In this section, we provide details on experimental settings, model implementation, and additional results.
\subsection{3D Visual Grounding}\label{supp:exp:grounding}
\paragraph{Setting} For all datasets, we evaluate all models with only the training sets provided. Following previous works~\cite{zhu20233d}, we report model performance on the validation set of all datasets in~\cref{tab:referit3d}. Notably, we used an off-the-shelf Mask3D segmentation model for generating object proposals with no optimization. 
\paragraph{Implementation} As briefly mentioned in~\cref{sec:exp:grounding}, we mainly considered three model settings in 3D visual grounding experiments, namely \textit{scratch}, \textit{pre-train}, and \textit{fine-tuned}. For the \textit{pre-train} setting, we follow the same setting mentioned in~\cref{sup:model:implement}. In the \textit{scratch} and \textit{fine-tuned} settings, to fairly compare with other dataset-specific fine-tuned models, we add an additional 2-layer MLP over the object features from the referral grounding self-attention transformer. During training, we fine-tune this grounding head together with all model weights for 100 epochs with a learning rate of $1\times10^{-4}$ for the added projection layer and set all other settings the same as the implementation described in~\cref{sup:model:implement}.


\subsection{Zero-shot Transfer}\label{supp:exp:zero}
\paragraph{Setting} In the zero-shot experiments, we first construct the held-out test set by aggregating scene-text pairs in \dataset from scenes in ScanNet and MultiScan. Specifically, we use the validation set of ScanRefer, Nr3D, and Sr3D. For scene-text pairs in the \dataset-val, we construct the test set by randomly sampling $\frac{1}{5}$ of human-annotated object referrals in the MultiScan dataset. This results in a test set with around 1.7K object referrals randomly drawn from 8.5k human-annotated object referrals in the MultiScan dataset. In the \textit{zero-shot} settings, we use all scene-text pairs from datasets in \dataset except for ScanNet and MultiScan. This includes both human-annotated and generated texts in ARKitScenes, 3RScan, and HM3D. This setting serves to test models' generalization capability in grounding objects with both unseen scenes and unseen texts. In the \textit{zero-shot text} setting, we add generated scene-text pairs in ScanNet and MultiScan into the data used in the \textit{zero-shot} setting, thereby making the held-out test contain mainly unseen object referrals.

\paragraph{Implementation} In the zero-shot experiments, we mainly considered three model settings \textit{scratch}, \textit{zero-shot}, and \textit{zero-shot text}. For the \textit{zero-shot} setting, we pre-train the model following~\cref{sup:model:implement} without additional grounding heads considering there is no additional training data available in the zero-shot transfer setting. In the \textit{scratch} and \textit{zero-shot text} setting, we follow the model implementation described in~\cref{supp:exp:grounding} and add an additional 2-layer MLP over the object features from the self-attention transformer. We follow the same fine-tuning setting described in~\cref{supp:exp:grounding}.

\subsection{3D question answering}
\paragraph{Setting} In the \ac{3dqa} experiments, we evaluate all models with only the training sets provided for fine-tuning. Following previous works~\cite{zhu20233d}, we report model performance on the validation and test sets of ScanQA and the test set of SQA3D.
\paragraph{Implementation} As briefly mentioned in~\cref{sec:exp:qa_seg}, we mainly considered the fine-tuned \model when comparing with existing methods. For all datasets, we initialize our \model model from a checkpoint pre-trained with 3D visual grounding on \dataset. We follow 3D-VisTA and add an additional question-answering module over the pre-trained representations for the answer prediction. We fine-tune the model for 100 epochs with a learning rate of $1\times 10^{-4}$ for the added question answering head and set all other settings the same as the implementation described in~\cref{sup:model:implement}.

\subsection{Open-vocabulary 3D semantic segmentation} 
\paragraph{Setting} 
Following RegionPLC~\cite{yang2023regionplc} proposed by Yang \etal, we conduct experiments to assess the performance of \dataset on open-vocabulary 3D semantic segmentation (OV-Seg). To establish a benchmark for open-vocabulary semantic segmentation, we adopt the experimental setup outlined in PLA~\cite{ding2023pla}, as per the methodology of RegionPLC. We utilize the annotation-free training setting, as described in RegionPLC, wherein semantic labels for all categories are omitted. This approach allows us to evaluate the effectiveness of \dataset in facilitating open-vocabulary segmentation without relying on predefined semantic segmentation annotations. For evaluation, we compute the mean Intersection over Union (mIoU) and mean accuracy (mAcc) across 17 foreground categories, excluding ``wall'' and ``floor'' background classes, as well as the ``other furniture'' category due to its inherent ambiguity.
\paragraph{Implementation}
We employ SparseUNet~\cite{graham20183d} as our 3D backbone network for extracting point features. We utilize different variants of SparseUNet, varying the number of channels in the input layer to explore its impact on performance. For text feature extraction, we employ CLIP~\cite{hegde2023clip} text encoder. To align the extracted features from the 3D scene encoder and the text encoder, we incorporate a vision-language adapter. The only supervision comes from the point-discriminative contrastive loss proposed by RegionPLC. During training, we employ the AdamW optimizer to update model parameters. We train the model from scratch for 500 epochs, utilizing a learning rate of $1\times 10^{-3}$. Additionally, we incorporate a warm-up period of 200 steps and a cosine annealing scheme for learning rate scheduling, with a minimum learning rate ratio of $1\times 10^{-5}$. 

\section{Additional Results}\label{supp:additional}

\subsection{Semantic Segmentation}\label{supp:exp:additional:semantic}
\paragraph{Setting} To test if the scaling effect of \dataset is universally beneficial for 3D understanding tasks, we use 3D semantic segmentation as a signature task to illustrate the effectiveness of \dataset. Notably, a recent work that introduced the Swin3D model~\cite{yang2023swin3d} has identified the importance of pre-training for 3D semantic segmentation~\cite{yang2023swin3d}. Following the same setting, we test if the proposed \textsc{Swin}3D model could be further improved by substituting the pre-training data to \dataset. Specifically, we test models' performance on the ScanNet semantic segmentation task with 20 semantic categories and report the mean IoU and mean Acc on the validation set of ScanNet. As the original implementation of \textsc{Swin}3D pre-training requires surface normals as additional inputs, we reimplement the model and pre-train all models with only point coordinates and colors.

\begin{table}
    \caption{\textbf{Semantic segmentation results on ScanNet validation set.} $\dagger$ denotes model trained with surface normals as an additional input. S3D indicates models initialized with the original \textsc{Swin}3D model weights pre-trained on Structured3D provided by Yang \etal~\cite{yang2023swin3d}.}
    \centering
    \resizebox{0.6\linewidth}{!}{
        \begin{tabular}{lcccc}
        \toprule
        Methods                                 & Init.            & \textsc{SceneVerse} Pre.    & mIoU                 & mAcc \\ \hline
        \textsc{Swin}3D$_n$-S$\dagger$                   & \xmark          & \xmark                  & 75.2                 & -                     \\
        \textsc{Swin}3D$_n$-S$\dagger$                   & S3D          & \xmark                  & 75.6                 & -                     \\ \hline
        \textsc{Swin}3D-S                      & \xmark          & \xmark                  & 63.2                 & 72.8                  \\
        \textsc{Swin}3D-S                      & S3D          & \xmark                  & 64.1                 & 75.1                  \\ 
        \textsc{Swin}3D-S~(\textit{pre-train})   & \xmark         & \cmark                  & 67.7                 & 78.0                  \\
        \textsc{Swin}3D-S~(\textit{pre-train})   & S3D         & \cmark                  & 69.5                 & 80.1                  \\
        \textsc{Swin}3D-S~(\textit{fine-tuned})  & S3D          & \cmark                  & \textbf{70.6}        & \textbf{80.2}         \\ \bottomrule
        \end{tabular}
    }
    \label{tab:semantic_seg}
\end{table}


\paragraph{Comparison} 
As shown in~\cref{tab:semantic_seg}, we observe a significant model performance improvement ($\sim$6\%) by training \textsc{Swin}3D-S model on our \dataset dataset. Comparing our pre-training set to Structured 3D, we also observe consistent model performance improvement, showcasing the benefit of scaling-effect in \dataset.
Moreover, we fine-tune the model on ScanNet after pre-training on \textsc{SceneVerse}. This process further brings improvement in model performance on semantic segmentation. We believe these results serve as strong pieces of evidence validating the effectiveness of data scaling in \dataset and also its potential benefit for all 3D tasks in addition to 3D visual grounding.



\subsection{Qualitative Results}\label{supp:exp:additional:qualitative}
\begin{figure*}[htbp!]
    \centering
    \includegraphics[width=0.85\linewidth]{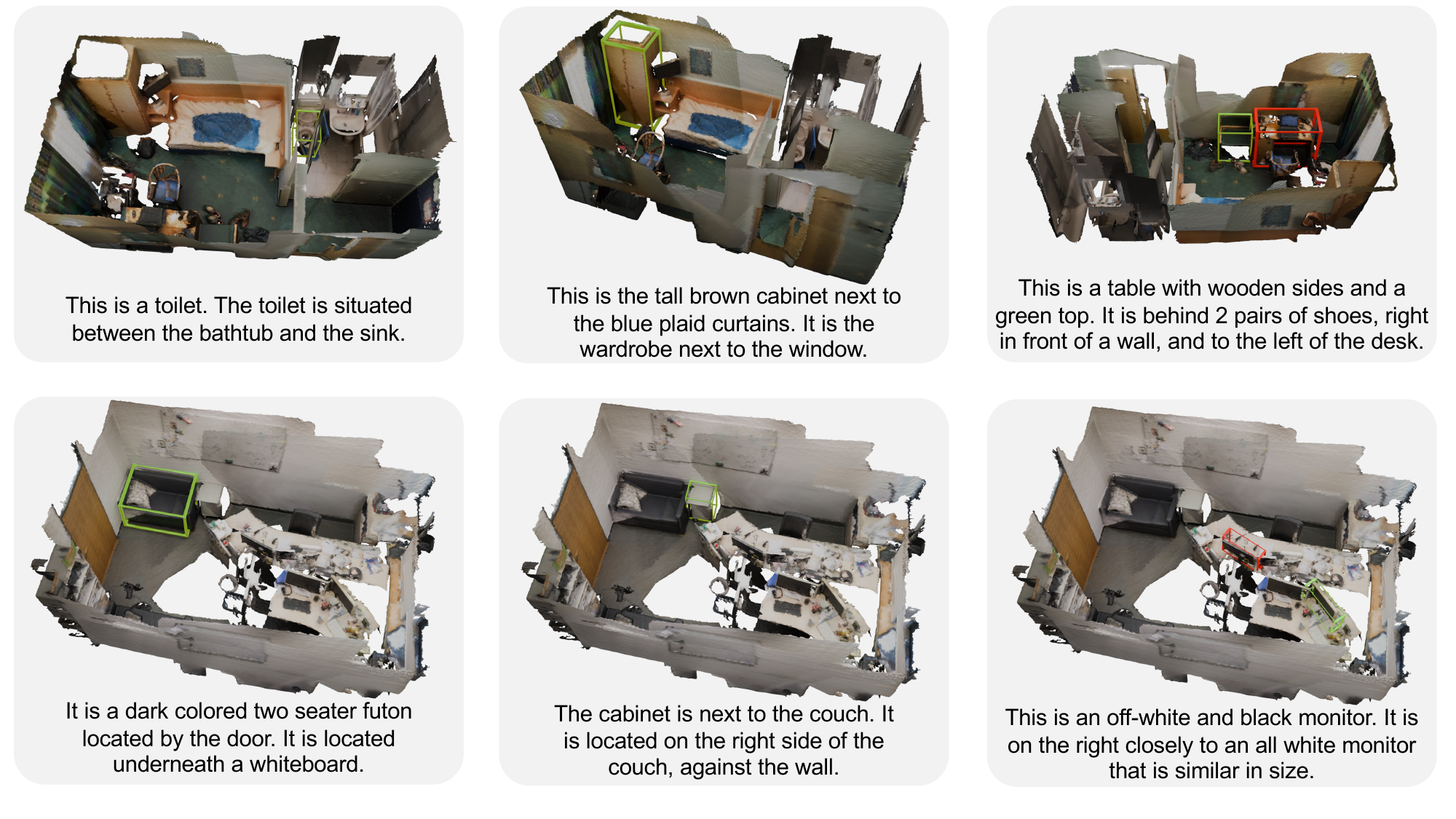}
    \caption{\textbf{Qualitative results of GPS on 3D visual-language grounding.}  We visualize the incorrect predictions in red and the correct predictions or ground truths in green.}
    \label{fig:refer_qualitative}
\end{figure*}

\begin{figure*}[htbp!]
    \centering
    \includegraphics[width=0.85\linewidth]{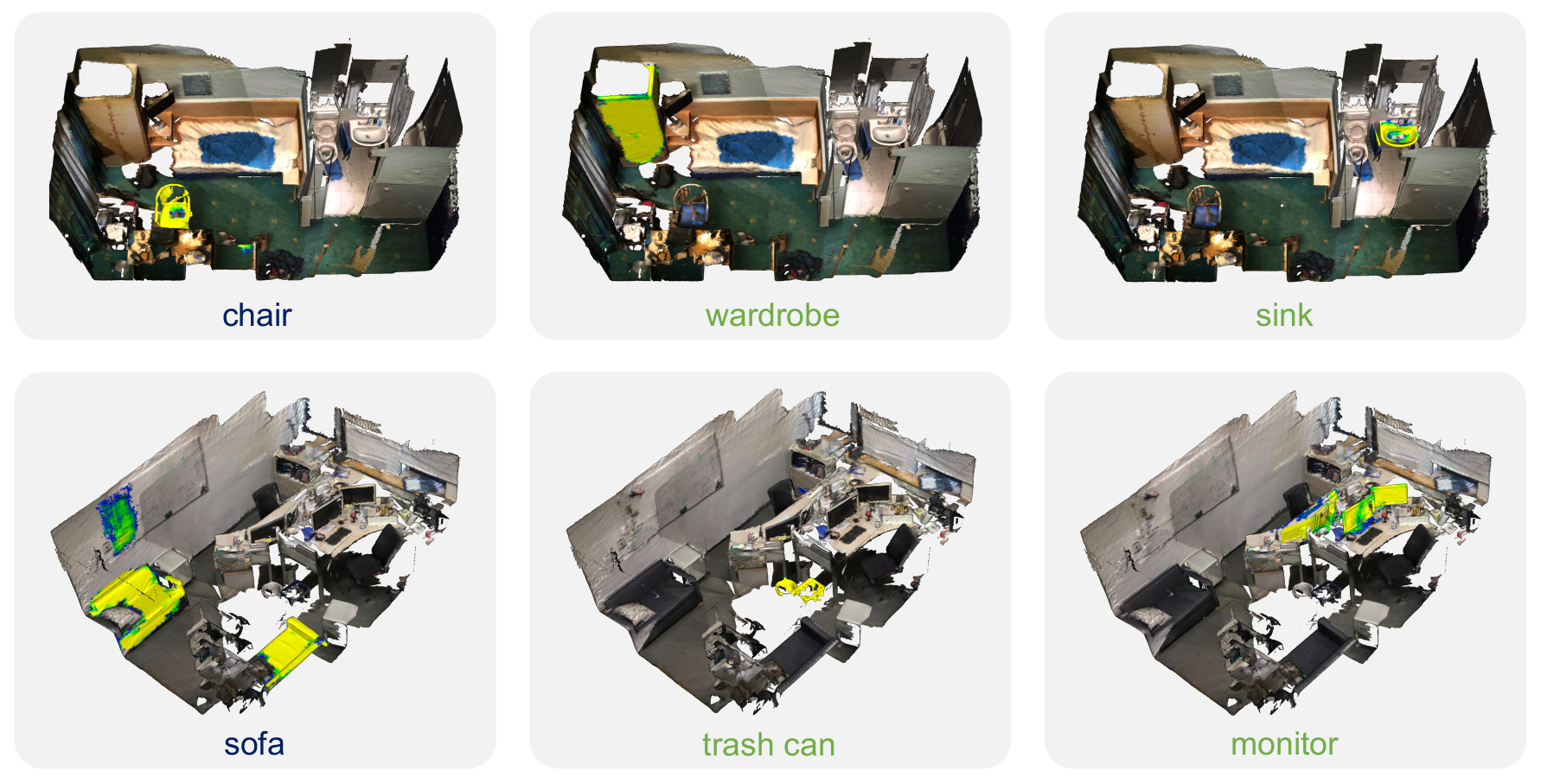}
    \caption{\textbf{Qualitative results of open-vocabulary 3D semantic segmentation (OV-Seg).}  We label the object class in ScanNet-20's vocabulary in blue, and unseen class in ScanNet-20 in green.}
    \label{fig:open_vocab_sem_seg_vis}
\end{figure*}

We provide the qualitative results of 3D vision-language grounding in~\cref{fig:refer_qualitative} and the results of open-vocabulary semantic segmentation in~\cref{fig:open_vocab_sem_seg_vis}.

\end{document}